\documentclass[10pt,journal,compsoc]{IEEEtran}

\ifCLASSINFOpdf

\else
\fi

\usepackage{amssymb}

\usepackage{amsmath}

\usepackage{booktabs}

\usepackage{multirow}
\usepackage{color}
\usepackage[top=2cm, bottom=2cm, left=2cm, right=2cm]{geometry}
\usepackage[linesnumbered,ruled,lined]{algorithm2e}
\usepackage{algorithmicx}
\usepackage{algpseudocode}
\usepackage{amsmath}

\usepackage{flushend}
\usepackage{diagbox}

\usepackage{threeparttable}

\usepackage{graphicx}
\usepackage{epstopdf}
\usepackage{subfigure}

\usepackage{CJK}

\usepackage{array}

\usepackage{stfloats}
\usepackage{amsthm}

\begin{document}

\title{Hybrid Local Causal Discovery}

\author{Zhaolong Ling,
        Honghui Peng,
        Yiwen Zhang*,
        Debo Cheng, \\
        Xingyu Wu,
        Peng Zhou,
        and Kui Yu
     \IEEEcompsocitemizethanks{

  	\IEEEcompsocthanksitem Z. Ling, H. Peng, Y. Zhang, and P. Zhou are with the School of Computer Science and Technology, Anhui University, Hefei, Anhui, 230601, China. Email: zlling@ahu.edu.cn, penghonghui@stu.ahu.edu.cn, zhangyiwen@ahu.edu.cn, and zhoupeng@ahu.edu.cn.  (*Corresponding author: Yiwen Zhang)
  	
  	\IEEEcompsocthanksitem D. Cheng is with the University of South Australia. E-mail: debo.cheng@unisa.edu.au.
  	
  	\IEEEcompsocthanksitem X. Wu is with the School of Hong Kong Polytechnic University, Department of Computing, Hong Kong, 999077, China. E-mail: xingy.wu@polyu.edu.hk.
  	
  	\IEEEcompsocthanksitem K. Yu is with the School of Computer Science and Information Technology, Hefei University of Technology, Hefei, 230009, China. E-mail: yukui@hfut.edu.cn.
  	
}

}


\markboth{IEEE TRANSACTIONS ON KNOWLEDGE AND DATA ENGINEERING ,~Vol.~14, No.~8, June~2015}
{Shell \MakeLowercase{\textit{et al.}}: Bare Demo of IEEEtran.cls for IEEE Journals}

\IEEEtitleabstractindextext{
	
\begin{abstract}

Local causal discovery aims to identify and distinguish the direct causes and effects of a target variable from observational data. Due to the inherent incompleteness of local information, popular methods from global causal discovery often face new challenges in local causal discovery tasks, such as 1) erroneous symmetry constraint tests and the resulting cascading errors in constraint-based methods, and 2) confusion within score-based approaches caused by local spurious equivalence classes. To address the above issues, we propose a \underline{H}ybrid \underline{L}ocal \underline{C}ausal \underline{D}iscovery algorithm, called HLCD. Specifically, HLCD initially utilizes a constraint-based approach with the OR rule to obtain a candidate skeleton, which is subsequently refined using a score-based method to eliminate redundant structures. Furthermore, during the local causal orientation phase, HLCD distinguishes between V-structures and equivalence classes by comparing local structure scores between the two, thereby avoiding orientation interference caused by local equivalence class ambiguities.  Experiments on 14 benchmark Bayesian networks and two real datasets validate that the proposed algorithm outperforms the existing local causal discovery methods.

\end{abstract}

 }

\maketitle

\IEEEdisplaynontitleabstractindextext

\IEEEpeerreviewmaketitle


\IEEEraisesectionheading{\section{Introduction}}

\IEEEPARstart{C}{ausal}  discovery has always been an important goal in many areas of scientific research~\cite{huang2019causal,prosperi2020causal}. It reveals the underlying causal mechanisms of data generation and contributes to solving decision-making problems in machine learning ~\cite{yu2020causality,chen2023some,Ling2025LabelAware}. Learning a Bayesian network (BN) from observational data is a popular method for causal discovery~\cite{cui2020causal,zhang2021crowd}. The structure of a BN takes the form of a directed acyclic graph (DAG), where nodes signify variables, and edges represent cause-effect relationships between variables~\cite{xie2020generalized,spirtes2000causation}.
In recent years, many global causal discovery algorithms have been proposed, which aim to learn the entire causal network~\cite{chickering2004large}. In general, learning a global causal network over a large number of variables is computationally intractable~\cite{zeng2021nonlinear,scutari2019learning}. To reduce computational complexity, the local-to-global approach was introduced to limit the search space for causal networks~\cite{tsamardinos2006max,gao2017local,guo2023adaptive}. Rather than exploring the entire network across all variables at once, these methods first identify the Markov blanket (MB) or the parent-child (PC) set of a target variable and gradually construct the DAG skeleton from these subsets~\cite{yu2023feature}.  In many practical cases, however, focusing on the causal relationships around a specific variable can eliminate the need to build a global causal network~\cite{wu2023feature}, increasing the importance of local causal discovery algorithms.


\begin{figure*}[t]
	\centering
	\includegraphics[height=3.10in]{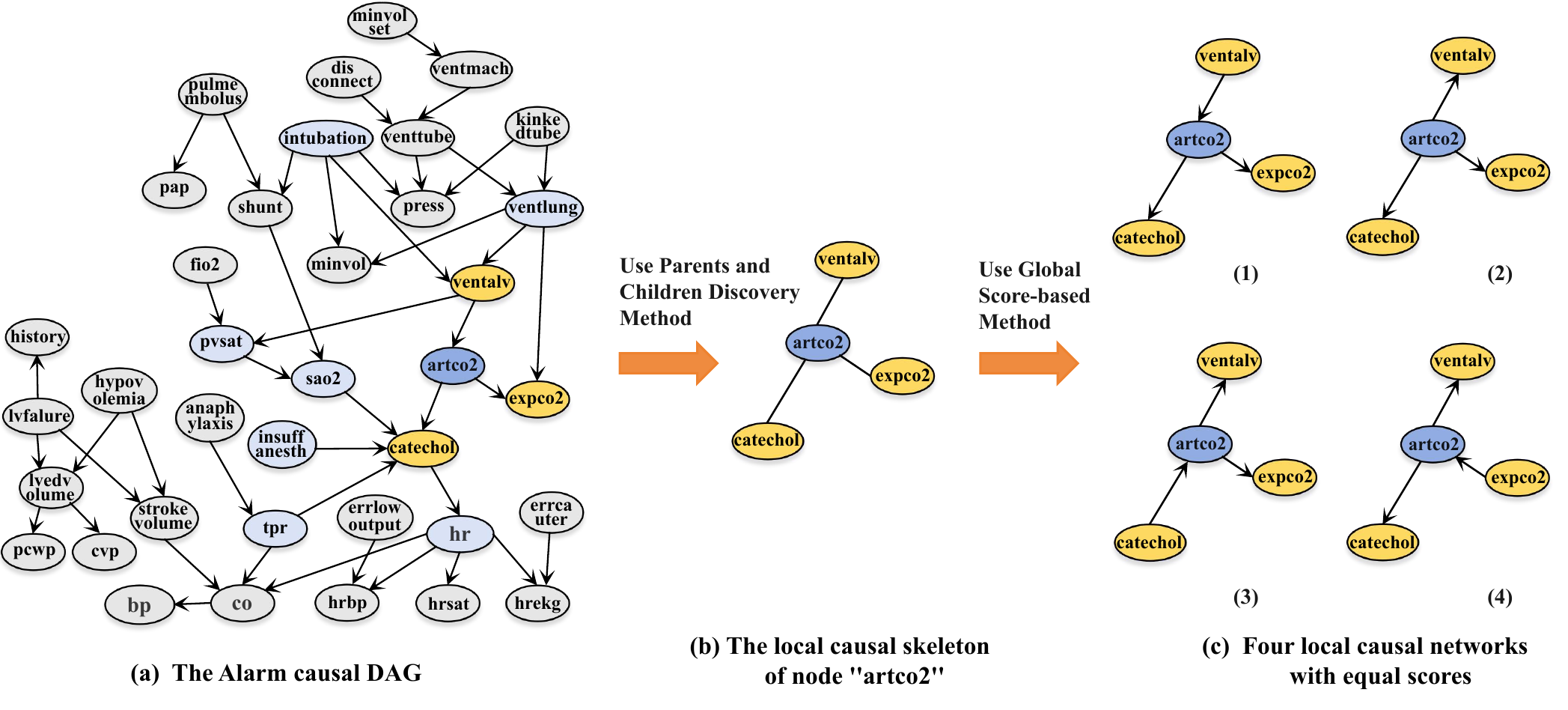}
	\caption{Directly using the search scoring algorithm to find the maximum score local network structure of node ``artco2'' will randomly return one of the four local structures in (c).  It may depend on the order in which the variables in the dataset are encountered.}
\end{figure*}

Local causal discovery aims to uncover the causal structure surrounding a specific variable. However, due to the unavailability of complete global information, many edge directions determined by relationships with distant variables\footnote{``Distant variables'' refer to nodes that are located further away from the target variable along the causal paths, i.e., paths involving multiple intermediate nodes in a DAG.} remain unidentified. As a result, most existing methods adopt a progressive learning approach to gradually acquire outer layer information, until the causal directions around the target variable are identified. Consequently, local causal discovery commonly employs the faster constraint-based methods~\cite{you2023local,ling2025local}, as score-based methods exhibit higher time complexity and are not well-suited for this gradual information acquisition process~\cite{ling2022light}.


Similar to global causal discovery methods, local causal discovery is susceptible to common issues associated with conditional independence (CI) testing, which can impact accuracy\cite{wu2024nonlinear}. One prominent concern is that CI testing cannot accurately determine the causal skeleton. As a result, many approaches employ symmetry tests to address this limitation~\cite{guo2022causal,wu2022multi}.  However, the prevailing AND rule\footnote{If $X$ belongs to the PC of $Y$, but $Y$ does not belong to the PC of $X$, then $X$ should be removed from the PC of $Y$} and OR rule\footnote{If $X$ belongs to the PC of $Y$, but $Y$ does not belong to the PC of $X$, then $Y$ should be added to the PC of $X$} used in these symmetry tests introduce certain errors. The AND rule aims to rigorously eliminate all erroneous relationships, while the OR rule  seeks to include as many true positives as possible, operating under a more lenient criterion. Empirical studies have provided evidence that approaches based on the AND rule achieve superior precision, whereas methods based on the OR rule exhibit better recall \cite{wu2021separation}. Consequently, neither approach yields completely accurate results. Additionally, the presence of data bias caused by inherent noise and incomplete local information further compounds the negative impact, exacerbating the potential for misleading results in local causal discovery.

To battle the challenge of insufficient observational samples in rare, costly, or privacy-sensitive events and the absence of global information due to unknown or unconsidered  distant causal relationships in local causal discovery, a natural approach is to leverage a hybrid methodology for local causal discovery. This approach seeks to enhance performance by combining the strengths of constraint-based and score-based methods.
While hybrid methods are commonly employed in global causal discovery research, their application in local causal discovery remains relatively unexplored. The complexity arises because a straightforward combination of these two methods inevitably faces the efficiency dilemma mentioned earlier in score-based approaches. Moreover, directly utilizing a global search scoring method to find the maximum score of local network structures may lead to incorrect local causal networks due to local equivalence class issues, as shown in Fig. 1. Furthermore, inaccuracies in CI tests resulting from information miss cascade into errors in the score-based causal discovery process. Consequently, effectively leveraging score information in local causal discovery poses a significant challenge.

In this paper, we introduce a novel hybrid method that identifies causal skeletons and V-structures by comparing scores among different local causal structures.  Specifically, we employ a constraint-based approach for the initial causal skeleton, which uses symmetric tests with the OR rule to achieve a comprehensive yet less precise structure. On this basis, we demonstrate the identification and removal of redundant structures through specialized local structure scores between the target variable and its causes and effects. Additionally, we prove the discovery of V-structures using similar score information. Our main contributions are summarized as follows:

\begin{itemize}
	
	\item  We theoretically analyze the special local structure score relationships between the target variable and its causal variables,  as well as different local structure score relationships between equivalence classes and V-structures.

	\item  We propose a \underline{H}ybrid \underline{L}ocal \underline{C}ausal \underline{D}iscovery algorithm, HLCD. To the best of our knowledge, HLCD is the first work on hybrid local causal discovery. Based on our analysis, HLCD can effectively eliminate redundant causal skeletons and differentiate between V-structures and equivalence classes by scoring to avoid interference caused by score equivalence.

	\item We conducted extensive experiments against seven state-of-the-art local causal discovery algorithms on 14 benchmark BN datasets and two real datasets. The results show that our HLCD algorithm outperforms the compared methods, especially in the small sample case.

\end{itemize}

The remainder of this paper is organized as follows. Section 2 reviews the related work and Section 3 gives the notations and definitions. Section 4 describes the proposed HLCD algorithm in detail and Section 5 reports the experimental results. Section 6 summarizes the paper.


\section{Relate Work}

Most local causal discovery algorithms are constraint-based and rely on CI tests to build and orient causal networks. Notable early approaches, such as Local Causal Discovery (LCD)~\cite{cooper1997simple} and its variants, use CI tests to discover causal relationships between sets of four variables. Bayesian Local Causal Discovery (BLCD) focuses on learning the $Y$-structure within the MB of a target variable~\cite{mani2004causal}. However, these LCD/BLCD algorithms aim to identify only a subset of causal edges, focusing on specific structural patterns among variables, without distinguishing the direct causal relationships for the target variable.

To address this problem, the state-of-the-art local causal discovery algorithms distinguish the direct causes and effects of the target variable directly. PCD-by-PCD (PCD means parents, children, and some descendants) \cite{yin2008partial} uses the  Max-Min Parents and Children (MMPC) algorithm \cite{tsamardinos2003time} to find PC and separating sets for V-structure identification, and then applies AND rule for local causal skeleton construction. Finally, the identified V-structures and Meek-rules~\cite{meek1995causal} are applied to orient the edges in the local causal skeleton. MB-by-MB~\cite{wang2014discovering} first finds a MB of the target node and constructs a local causal structure, and then sequentially finds MB of variables connected to the target and simultaneously constructs local structures along the paths starting from the target until the causes and effects of the target have been determined. Causal Markov Blanket (CMB)~\cite{gao2015local} initially applies the HITON-Markov Blanket (HITON-MB) algorithm~\cite{aliferis2003hiton} to identify the target's MB, and then orients the edges by monitoring changes in conditional independence within the MBs. The Local Causal Structure learning by Feature Selection algorithm (LCS-FS)~\cite{ling2020using} uses the mutual information-based feature selection method~\cite{peng2005feature} to discover the PC set of variables and construct the skeleton using OR rules, and then searches for separating sets from the learned PC sets and in turn uses the separating sets for edge orientation. Yang et al.~\cite{yang2021towards} proposed the concept of N-structures. By using N-structures, the Efficient Local Causal Structure Learning (ELCS) algorithm~\cite{yang2021towards} uncovers the local structure of the target variable while minimizing the number of MBs learned, thus reducing the number and influence of unreliable CI tests. The Partial Structure Learning (PSL) algorithm~\cite{ling2022psl} is a partial causal discovery algorithm. It uses the OR rule to build the skeleton and finds two types of V-structures, Type-C and Type-NC, in the PC set of the current node, avoiding the false edge orientation problem of local causal discovery algorithms. Recently, Yang et al. proposed GraN-LCS (Gradient-based Local Causal Structure Learning)~\cite{liang2024gradient}, a gradient-based approach for learning local causal structures. This method builds a multilayer perceptron (MLP) to simultaneously model the relationships of all other variables with a target variable, and incorporates an acyclicity-constrained local recovery loss to encourage the discovery of local graphs and identify direct causes and effects.

However, existing local causal discovery algorithms do not consider utilizing score information in the data to enhance the performance of the algorithm when faced with challenges such as sample size, noise, and global information deficiencies. In this paper, we will develop a novel hybrid local discovery algorithm to improve the accuracy of local causal discovery.

\section{Notations And Definitions}

In this section, we will briefly introduce some basic notations and definitions. 

\noindent \textbf{Definition 1 (Bayesian Network)}~\cite{spirtes2000causation}. The Bayesian Network, ${\mathbb{B}}(\mathcal{G},\Theta )$, is defined by a tuple consisting of a DAG $\mathcal{G}$, and a set of parameters $\Theta$.  The graph $\mathcal{G}$ consists of a set of nodes, \textbf{U}, and a set of directed edges, \textbf{E}, where nodes denote variables and edges denote causal relationships between variables.  The parameters $\Theta$ are the specific quantification of the strength of the dependencies between nodes.

When the variables in a BN are discrete, the parameters $\Theta$ represent the conditional probabilities of the variables, and the BN model is parameterized by a conditional probability tables (CPTs).  When the variables in a BN are continuous, the network model is parameterized with conditional probability distributions (CPDs). It is important to note in particular that our BN model follows variables in discrete form, and the model is parameterized by a CPTs, in which the parameters $\Theta$ are the conditional probabilities of the variables under their parent nodes.

\noindent \textbf{Definition 2 (V-Structure)}~\cite{spirtes2000causation}. The triplet of variables $X$, $Y$, and $T$ forms a V-structure and $T$ is a collider if node $T$ has two incoming edges from $X$ and $Y$ respectively, i.e. $X \rightarrow T \leftarrow Y$, and $X$ is not adjacent to $Y$.

\noindent \textbf{Definition 3 (Symmetry Constraint)}~\cite{gao2017efficient}. For a node $X$ to be a parent or child of $Y$ in a DAG. Then, $X$ must be in the PC set of $Y$ and $Y$ must be in the PC set of  $X$, i.e., $X \in \textbf{PC}_{Y}$ and $Y \in \textbf{PC}_{X}$.

In a BN, learning causal relationships between variables from data is sometimes asymmetric. In this case, to satisfy the symmetry constraint, the AND and OR rules are usually used as constraint tests to ensure symmetry~\cite{guo2022causal}.

\noindent \textbf{Definition 4 (Score Consistency)}~\cite{chickering2002optimal}. Let $\mathcal{D}$ be a set of data consisting of i.i.d. samples from some distribution ${\mathcal{P}}$. A score criterion $\mathcal{S}$ is \textbf{consistent} if, as the size of $\mathcal{D}$ goes to infinity, the following two properties hold true: 

1) if the structure $\mathcal{G}$ contains ${\mathcal{P}}$ and another structure $\mathcal{G}'$ does not, then $\mathcal{S}(\mathcal{G},\mathcal{D})>\mathcal{S}(\mathcal{G}',\mathcal{D})$;

2) if $\mathcal{G}$ and $\mathcal{G}'$ both contain  ${\mathcal{P}}$ but $\mathcal{G}$ has fewer parameters, then $\mathcal{S}(\mathcal{G},\mathcal{D})>\mathcal{S}(\mathcal{G}',\mathcal{D})$.

The graph $\mathcal{G}$ contains ${\mathcal{P}}$ if there exists a set of parameter values $\Theta$ for $\mathcal{G}$ such that the parameterized BN model $(\mathcal{G},\Theta)$ represents ${\mathcal{P}}$ exactly.

\noindent \textbf{Definition 5 (Local Score Consistency)}~\cite{chickering2002optimal}. Let $\mathcal{D}$ be a set of data consisting of i.i.d. samples from some distribution  ${\mathcal{P}}$. Let $\mathcal{G}$ be any BN structure and $\mathcal{G}'$ be the same structure as $\mathcal{G}$ but with an edge from a node $Y$ to a node $X$. Let $\textbf{Pa}_{X}^{\mathcal{G}}$ be the parent set of $X$ in $\mathcal{G}$. A score criterion $\mathcal{S}$ is \textbf{locally consistent} if, as the size of $\mathcal{D}$ goes to infinity, the following two properties hold true: 

1) if $X \not\!\perp\!\!\!\perp Y\mid \textbf{Pa}_{X}^{\mathcal{G}}$, then $ \mathcal{S}(\mathcal{G},\mathcal{D})<\mathcal{S}(\mathcal{G}',\mathcal{D})$;

2) if $X \!\perp\!\!\!\perp Y\mid \textbf{Pa}_{X}^{\mathcal{G}}$, then $ \mathcal{S}(\mathcal{G},\mathcal{D})>\mathcal{S}(\mathcal{G}',\mathcal{D})$. 

The score-based methods rely on score criteria to learn the best-fit DAG $\mathcal{G}$ for the data samples. In general, the higher the score for $\mathcal{G}$, the better the fit to the data $\mathcal{D}$, and vice versa.

For example, information theoretic scores~\cite{kitson2023survey} aim to avoid over-fitting by balancing the goodness of fit with model dimensionality given the available data. The general form of these scores can be expressed as:
\begin{equation}
	\mathcal{S}(\mathcal{G},\mathcal{D}) = \log [\hat p(\mathcal{D}|\mathcal{G})] - \Delta (\mathcal{D},\mathcal{G})
\end{equation}

\noindent where $\log \hat p(\mathcal{D}|\mathcal{G})$ denotes the goodness of fit as measured by the log likelihood of the data given the graph, in the case where the distribution parameters, $\Theta$, take their maximum likelihood estimation values. The $\Delta (\mathcal{D},\mathcal{G})$  is a function which penalises graph complexity. One of the most commonly used information theoretic scores is the AIC score function (denoted as 	${\mathcal{S_A}}(\mathcal{G},\mathcal{D})$), calculated as follows
\begin{equation}
	{\mathcal{S_A}}(\mathcal{G},\mathcal{D}) = \sum\limits_{i = 1}^n {\sum\limits_{j = 1}^{{q_i}} {\sum\limits_{k = 1}^{{r_i}} {{N_{ijk}}\log \frac{{{N_{ijk}}}}{{{N_{ij}}}}} } }-\sum\limits_{i = 1}^n {({r_i} - 1){q_i}} 
\end{equation}
In the definition of the above formula, $i$ is the index over the $n$ variables, $j$ is the index over the $q_i$ combinations of values of the parents of the node $X_i$, and $k$ is the index over the $r_i$ possible values  of node $X_i$.  $N$ is the sample size and ${N_{ij}} = \sum\nolimits_{k = 1}^{{r_i}} {{N_{ijk}}}$.

Bayesian scores~\cite{kitson2023survey} consider prior beliefs about the graphical structure and dependence relationship parameters to select the network structure with the largest posterior probability in the data.
\begin{equation}
	\mathcal{S}(\mathcal{G},\mathcal{D}) = arg\max p(\mathcal{G}|\mathcal{D}) = arg\max \frac{{p(\mathcal{D}|\mathcal{G})p(\mathcal{G})}}{{p(\mathcal{D})}}
\end{equation}
In general, one assumes that all graph structures are equally probable a priori, and the most commonly used Bayesian score is the BDeu score function (denoted as ${\mathcal{S}_\mathcal{B}}(\mathcal{G},\mathcal{D})$) where the prior parameters for $\theta {_{ijk}}$ are set to $1/{r_i}{q_i}$ for all $i$, $j$, $k$ leading to the following definition:
\begin{equation}
	\begin{aligned}
		{\mathcal{S}_\mathcal{B}}(\mathcal{G},\mathcal{D}) = \log p(\mathcal{G}) +& \sum\limits_{i = 1}^n {\sum\limits_{j = 1}^{{q_i}} { [\log \frac{{\Gamma (N'/{q_i})}}{{\Gamma ({N_{ij}} + N'/{q_i})}}} }   \\
		& + \sum\limits_{k = 1}^{{r_i}} {\log \frac{{\Gamma ({N_{ijk}} + N'/{r_i}{q_i})}}{{\Gamma (N'/{r_i}{q_i})}}} ]
	\end{aligned}
\end{equation}
where $p(\mathcal{G}|\mathcal{D})$ denotes the conditional probability of graph $\mathcal{G}$ under data $\mathcal{D}$, $p(\mathcal{G})$ is the prior probability of a particular graph structure which is assumed to be the same for all graphs, $\Gamma$ is the Gamma function, and $N'$ is the equivalent sample size which expresses our confidence in the prior parameters.

\section{The Proposed Method}

In this section, we provide a detailed introduction to our method, including theoretical analysis and proofs, implementation details, and time complexity analysis.

\subsection{The local causal discovery strategy}

In this section, we introduce the hybrid local causal discovery strategy for HLCD. This strategy is constructed based on the following two fundamental theorems. 

Before introducing and proving the theorem, we need to account for the symbolic representation. Since the AIC score function (denoted as 	${\mathcal{S_A}}(\mathcal{G},\mathcal{D})$)  and BDue score function (denoted as ${\mathcal{S}_\mathcal{B}}(\mathcal{G},\mathcal{D})$)  are decomposable, they can be written as a sum of metrics, each of which is a function of only one node and its parent node (i.e. ${\mathcal{S}_{\mathcal{A}/\mathcal{B}}}(\mathcal{G},\mathcal{D}) = \sum\nolimits_{i = 1}^n {{\mathcal{S}_{\mathcal{A}/\mathcal{B}}}({X_i},Pa_i^\mathcal{G})} $, where $Pa_i^\mathcal{G}$ denotes the parent of $X_i$ in $\mathcal{G}$). Therefore, we use the symbols ${\mathcal{S}_{\mathcal{A}/\mathcal{B}}}(\emptyset  \rightarrow X,\mathcal{D})$ and ${\mathcal{S}_{\mathcal{A}/\mathcal{B}}}(X  \rightarrow Y,\mathcal{D})$ to denote the AIC score or BDue score  of node $X$ and $Y$, respectively. Where  ${\mathcal{S}_{\mathcal{A}/\mathcal{B}}}(\emptyset  \rightarrow X,\mathcal{D})$ denotes the score of node $X$ when the empty set is the parent of $X$, and ${\mathcal{S}_{\mathcal{A}/\mathcal{B}}}(X  \rightarrow Y,\mathcal{D})$ denotes the score of node $Y$ when $X$ is the parent of $Y$.  Moreover,  \textbf{U} denotes the set of variables in the dataset, and $ \textbf{PC}_{T} $ refers to the parents and children nodes of the target variable $T$.

\textbf{Theorem 1.} Let $T$ be any variable in \textbf{U},  and $X$ be a variable in $ \textbf{PC}_{T} $. Assume that the score function maintains local score consistency within the data $\mathcal{D}$. When node $X$ is treated as a parent of $T$, in the local structure \( X \rightarrow T \), the score of node $T$ will increase. Conversely, when node $T$ is treated as a parent of $X$, in the local structure \( T \rightarrow X \), the score of node $X$ will increase. Moreover, the score gains in both cases are identical. i.e. $ {\mathcal{S}_{\mathcal{A}/\mathcal{B}}}(X  \rightarrow T,\mathcal{D}) - {\mathcal{S}_{\mathcal{A}/\mathcal{B}}}(\emptyset  \rightarrow T,\mathcal{D})   = {\mathcal{S}_{\mathcal{A}/\mathcal{B}}}(T \rightarrow X,\mathcal{D}) - {\mathcal{S}_{\mathcal{A}/\mathcal{B}}}(\emptyset  \rightarrow X,\mathcal{D}) > 0$  holds.

\emph{Proof:} Assuming that the local structure $X \rightarrow T$ is denoted as $\mathcal{G}$, and $T \rightarrow X$ is denoted $\mathcal{G}'$. The AIC score for $\mathcal{G}$ is calculated as follows:
\begin{equation}
	\begin{aligned}
		{\mathcal{S_A}}(\mathcal{G},\mathcal{D}) = & {\mathcal{S_A}}(\emptyset  \rightarrow X,\mathcal{D}) + {\mathcal{S_A}}(X  \rightarrow T,\mathcal{D})
	\end{aligned}
\end{equation}
To calculate the maximum likelihood estimate of the parameter $\theta_{ijk}$ using ${N_{ijk}}\log ({N_{ijk}}/{N_{ij}})$, the above formula can be rewritten as follows:
\begin{equation}
	\begin{aligned}
		{\mathcal{S}_\mathcal{A}}(\mathcal{G},\mathcal{D}) =& \sum\limits_{k = 1}^{{r_X}} {{N_k}} \log \frac{{{N_k}}}{N}\\
		&+ \sum\limits_{j = 1}^{{q_T}} {\sum\limits_{k = 1}^{{r_T}} {{N_{jk}}} } \log \frac{{{N_{jk}}}}{{{N_j}}} - \Delta (\mathcal{D},\mathcal{G})
	\end{aligned}
\end{equation}
Writing the log-likelihood term in multiplicative form and using the conditional probability form to express $\frac{N_{ijk}}{N_{ij}}$ (i.e., for node $X$, its $\frac{N_{jk}}{N_{j}}$ can be denoted as $p(X = k \mid \text{Parents}(X) = j)$), we can further obtain the following equation:
\begin{equation}
	\begin{aligned}
		&	{\mathcal{S}_\mathcal{A}}(\mathcal{G},\mathcal{D})\\
		&= \log [\prod\limits_{{\text{k}} = 1}^{{r_X}} {p{{(X|\emptyset )}^{{N_k}}}} \prod\limits_{j = 1}^{{q_T}} {\prod\limits_{k = 1}^{{r_T}} {p{{(T|X)}^{{N_{jk}}}}]} } - \Delta (\mathcal{D},\mathcal{G})
	\end{aligned}
\end{equation}
Writing ${p(T|X)}$ in the form of the posterior probability ${p(X|T) p(T)/p(X)}$. In addition, it is important to note that the number of terms in the log-likelihood of the local scores of each node is the size of  $\mathcal{D}$, so the following equation holds:
\begin{equation}
	\begin{aligned}
		&{\mathcal{S}_\mathcal{A}}(\mathcal{G},\mathcal{D})\\
		&=\log [\prod\limits_{{\text{k}} = 1}^{{r_X}} {p{{(X|\emptyset )}^{{N_k}}}} \prod\limits_{j = 1}^{{q_T}} {\prod\limits_{k = 1}^{{r_T}} {{{[\frac{{p(T)p(X|T)}}{{p(X)}}]}^{{N_{jk}}}}]} }  - \Delta (\mathcal{D},\mathcal{G})\\
		&=\log [\prod\limits_{k = 1}^{{r_T}} {p{{(T|\emptyset )}^{{N_k}}}} \prod\limits_{j = 1}^{{q_X}} {\prod\limits_{k = 1}^{{r_X}} {p{{(X|T)}^{{N_{jk}}}}} } ] - \Delta (\mathcal{D},\mathcal{G})\\
		&={\mathcal{S}_\mathcal{A}}(\emptyset  \rightarrow T,\mathcal{D}) + {\mathcal{S}_\mathcal{A}}(T  \rightarrow X,\mathcal{D}) + \Delta (\mathcal{D},\mathcal{G'}) - \Delta (\mathcal{D},\mathcal{G})
	\end{aligned}
\end{equation}

Since local structure $X \rightarrow T$ and $ T \rightarrow X$ have the same penalty term (i.e. $\Delta (\mathcal{D},\mathcal{G}) = \Delta (\mathcal{D},\mathcal{G'}) = {r_X}{r_T} - 1$), it follows from above that $ {\mathcal{S}_\mathcal{A}}(X  \rightarrow T,\mathcal{D}) - {\mathcal{S}_\mathcal{A}}(\emptyset   \rightarrow T,\mathcal{D})   = {\mathcal{S}_\mathcal{A}}(T \rightarrow X,\mathcal{D}) - {\mathcal{S}_\mathcal{A}}(\emptyset  \rightarrow X,\mathcal{D})$  holds.

For the BDeu score criterion, we also use $\mathcal{G}$ and $\mathcal{G'}$ to denote local structure $X \rightarrow T$ and  $T \rightarrow X$, respectively, then the BDeu score for $\mathcal{G}$ is calculated as follows:
\begin{equation}
	\begin{aligned}
		{\mathcal{S_B}}(\mathcal{G},\mathcal{D}) = & {\mathcal{S_B}}(\emptyset  \rightarrow X,\mathcal{D}) + {\mathcal{S_B}}(X  \rightarrow T,\mathcal{D})
	\end{aligned}
\end{equation}
Writing the prior parameter of $\theta {_{ijk}} $ in graph $\mathcal{G}$ in the form of $1/r_iq_i$, the following equation can be further obtained:
\begin{equation}
	\begin{aligned}
		&	{\mathcal{S}_\mathcal{B}}(\mathcal{G},\mathcal{D}) = \log p(\mathcal{G}) + \log \frac{{\Gamma (N')}}{{\Gamma (N + N')}} \\
		&+ \sum\limits_{k = 1}^{{r_X}} {\log \frac{{\Gamma ({N_k} + N'/{r_X})}}{{\Gamma (N'/{r_X})}}} + \sum\limits_{j = 1}^{{q_T}} {\log \frac{{\Gamma (N'/{q_T})}}{{\Gamma ({N_j} + N'/{q_T})}}} \\
		& + \sum\limits_{j = 1}^{{q_T}} {\sum\limits_{k = 1}^{{r_T}} {\log \frac{{\Gamma ({N_{jk}} + N'/{r_T}{q_T})}}{{\Gamma (N'/{r_T}{q_T})}}} } 
	\end{aligned}
\end{equation}
For structure $\mathcal{G}$, since the parent node of $T$ is $X$, then the equation $ q_T = r_X $ holds, and the number of samples for each state value of the parent node of $T$ is equal to the number of samples for each state value of node $X$, i.e., $ N_j = N_k$. In summary, we can conclude that equation $\sum\limits_{k = 1}^{{r_X}} {\log \frac{{\Gamma ({N_k} + N'/{r_X})}}{{\Gamma (N'/{r_X})}}}  =  - \sum\limits_{j = 1}^{{q_T}} {\log \frac{{\Gamma (N'/{q_T})}}{{\Gamma ({N_j} + N'/{q_T})}}}$ holds. Thus, the BDeu score for $\mathcal{G}$ can be written in the following form:
\begin{equation}
	\begin{aligned}
		{\mathcal{S}_\mathcal{B}}(\mathcal{G},\mathcal{D}) = & \log p(\mathcal{G}) + \log \frac{{\Gamma (N')}}{{\Gamma (N + N')}}\\
		& + \sum\limits_{j = 1}^{{q_T}} {\sum\limits_{k = 1}^{{r_T}} {\log \frac{{\Gamma ({N_{jk}} + N'/{r_T}{q_T})}}{{\Gamma (N'/{r_T}{q_T})}}} }
	\end{aligned}
\end{equation}
Similarly, for the graph $\mathcal{G'}$, the formula for its BDue can eventually be written in the following form:
\begin{equation}
	\begin{aligned}
		{\mathcal{S}_\mathcal{B}}(\mathcal{G'},\mathcal{D}) = & \log p(\mathcal{G'}) + \log \frac{{\Gamma (N')}}{{\Gamma (N + N')}}  \\
		&+ \sum\limits_{j = 1}^{{q_X}} {\sum\limits_{k = 1}^{{r_X}} {\log \frac{{\Gamma ({N_{jk}} + N'/{r_X}{q_X})}}{{\Gamma (N'/{r_X}{q_X})}}} }
	\end{aligned}
\end{equation}
Since BDeu assumes that all graph structures are equally probable a priori,  then $\log p(\mathcal{G}) = \log p(\mathcal{G'})$ hold. For structure $\mathcal{G}$, since the parent node of $T$ is $X$, then $ q_T = r_X $ holds. Similarly, for structure $\mathcal{G'}$, since the parent node of $X$ is $T$, then $q_X = r_T$ holds. Therefore, the formula  $r_T q_T = r_X q_X $ holds. As $N_{ij}$ represents the number of samples for each combination of state values of $X$ and $T$, in the structure $ X \rightarrow T $, for each $N_{ij}$ ($j \in {q_T}, k \in {r_T}$), there will be an equivalent $N_{ij}$ ($j \in {q_X}, k \in {r_X}$) in the structure $T \rightarrow X$, and the number of $N_{ij}$ terms will be equal, which is the number of combinations of state values of $X$ and $T$. In summary, $\sum\limits_{j = 1}^{{q_T}} {\sum\limits_{k = 1}^{{r_T}} {\log \frac{{\Gamma ({N_{jk}} + N'/{r_T}{q_T})}}{{\Gamma (N'/{r_T}{q_T})}}} }  = \sum\limits_{j = 1}^{{q_X}} {\sum\limits_{k = 1}^{{r_X}} {\log \frac{{\Gamma ({N_{jk}} + N'/{r_X}{q_X})}}{{\Gamma (N'/{r_X}{q_X})}}} } $ will hold. Therefore, the equation ${\mathcal{S}_\mathcal{B}}(X \to T,\mathcal{D}) -{\mathcal{S}_\mathcal{B}}(\emptyset  \to T,\mathcal{D}) = {\mathcal{S}_\mathcal{B}}(T \to X,\mathcal{D}) - {\mathcal{S}_\mathcal{B}}(\emptyset  \to X,\mathcal{D})$ holds.

On further analysis,  when $X \rightarrow T$ is the true local structure, the $X$ is the parent of $T$,  then $\forall \textbf{Z} \subseteq \textbf{U}\backslash \{X,T\}, X \not\! \perp\! \! \! \! \perp T\mid \textbf{Z} $ holds. Therefore, when the conditioning set is $\emptyset$, the  $X \not\!\perp\!\!\!\perp T\mid \emptyset $ holds. According to local consistency, the ${\mathcal{S}_{\mathcal{A}/\mathcal{B}}}(X \to T,\mathcal{D}) - {\mathcal{S}_{\mathcal{A}/\mathcal{B}}}(\emptyset  \to T,\mathcal{D})>0$ will hold. Similarly, when $T \rightarrow X$ is the true local structure, the inequality ${\mathcal{S}_{\mathcal{A}/\mathcal{B}}}(T \to X,\mathcal{D}) - {\mathcal{S}_{\mathcal{A}/\mathcal{B}}}(\emptyset  \to X,\mathcal{D})>0$ will hold. Based on the above conclusion, regardless of whether the true structure is $ X \rightarrow T $ or $T \rightarrow X$, the formula ${\mathcal{S}_{\mathcal{A}/\mathcal{B}}}(X \to T,\mathcal{D}) - {\mathcal{S}_{\mathcal{A}/\mathcal{B}}}(\emptyset  \to T,\mathcal{D}) = {\mathcal{S}_{\mathcal{A}/\mathcal{B}}}(T \to X,\mathcal{D}) - {\mathcal{S}_{\mathcal{A}/\mathcal{B}}}(\emptyset  \to X,\mathcal{D})$ holds. Therefore, when the true structure is $X \rightarrow T$, the node $T$ can increase the score of the local structure $\emptyset \rightarrow X$, and when the true structure is $T \rightarrow X $, node $X$ can also increase the score of $\emptyset \rightarrow T $.  According to the above analysis, we can ultimately establish that if $X \in \textbf{PC}_{T}$, then the conclusion $ {\mathcal{S}_{\mathcal{A}/\mathcal{B}}}(X  \rightarrow T,\mathcal{D}) - {\mathcal{S}_{\mathcal{A}/\mathcal{B}}}(\emptyset   \rightarrow T,\mathcal{D})   = {\mathcal{S}_{\mathcal{A}/\mathcal{B}}}(T \rightarrow X,\mathcal{D}) - {\mathcal{S}_{\mathcal{A}/\mathcal{B}}}(\emptyset  \rightarrow X,\mathcal{D}) > 0$ will holds. $\hfill\blacksquare$

Theorem 1 shows that when $X$ is a causal node of $T$, the local score relationship between $X$ and $T$ will always satisfy $ {\mathcal{S}_{\mathcal{A}/\mathcal{B}}}(X  \rightarrow T,\mathcal{D}) - {\mathcal{S}_{\mathcal{A}/\mathcal{B}}}(\emptyset   \rightarrow T,\mathcal{D})   = {\mathcal{S}_{\mathcal{A}/\mathcal{B}}}(T \rightarrow X,\mathcal{D}) - {\mathcal{S}_{\mathcal{A}/\mathcal{B}}}(\emptyset  \rightarrow X,\mathcal{D}) > 0$. Therefore, using Theorem 1, we can first employ existing parent and child discovery algorithms and the OR rule to construct a comprehensive but redundant causal skeleton. Then, the nodes in the skeleton that do not satisfy Theorem 1 are deleted, thus removing the redundant skeleton structure and providing a more precise causal structure search space for the subsequent causal orientation.

Since all structures within the equivalence class share the same score, we consider \( X \rightarrow T \rightarrow Y \) as the representative structure of the equivalence class, and \( X \rightarrow T \leftarrow Y \) as the V-structure.

\textbf{Theorem 2.} Let $X,Y,T \in \textbf{U}$ and $T$ be a target node with no edge connected between $X$ and $Y$,  and $ X,Y \in \textbf{PC}_{T} $. Assume that the score function maintains score consistency within the data $\mathcal{D}$.  Then, when the score of local structures $X \rightarrow T \leftarrow Y$ is greater than the score of local structures $X \rightarrow T \rightarrow Y$, i.e., ${\mathcal{S}_{\mathcal{A}/\mathcal{B}}}(\emptyset  \rightarrow X,\mathcal{D}) + {\mathcal{S}_{\mathcal{A}/\mathcal{B}}}(\emptyset  \rightarrow Y,\mathcal{D}) + {\mathcal{S}_{\mathcal{A}/\mathcal{B}}}(X,Y \rightarrow T,\mathcal{D}) > {\mathcal{S}_{\mathcal{A}/\mathcal{B}}}(\emptyset  \rightarrow X,\mathcal{D}) + {\mathcal{S}_{\mathcal{A}/\mathcal{B}}}(X \rightarrow T,\mathcal{D}) + {\mathcal{S}_{\mathcal{A}/\mathcal{B}}}(T \rightarrow Y,\mathcal{D})$, there exists a V-structure in variables $X$, $Y$, $T$, and $T$ is a collision node.

\emph{Proof.} Assuming that the local structure $X \rightarrow T \leftarrow Y$ is denoted as $\mathcal{G}$, and $X \rightarrow T \rightarrow Y$ is denoted $\mathcal{G}'$. The AIC score for ${\mathcal{S}_\mathcal{A}}(\mathcal{G},\mathcal{D}) - {\mathcal{S}_\mathcal{A}}(\mathcal{G}',\mathcal{D})$ is calculated as follows:
\begin{equation}
	\begin{aligned}
		{\mathcal{S}_\mathcal{A}}(\mathcal{G},\mathcal{D}) - {\mathcal{S}_\mathcal{A}}(\mathcal{G}',\mathcal{D})=&\log [\hat p(\mathcal{D}|\mathcal{G})]- \Delta (\mathcal{D},\mathcal{G})\\
		&- \log [\hat p(\mathcal{D}|\mathcal{G'})] + \Delta (\mathcal{D},\mathcal{G'})\\
	\end{aligned}
\end{equation}
Subtracting the same terms, the log-likelihood terms are written in multiplicative form and substituting the form of conditional probability for ${\frac{{{N_{ijk}}}}{{{N_{ij}}}}}$ (i.e., for node $X$, its $\frac{N_{jk}}{N_{j}}$ can be denoted as $p(X = k \mid \text{Parents}(X) = j)$),  the following equation can be obtained:
\begin{equation}
	\begin{aligned}
		&{\mathcal{S}_\mathcal{A}}(\mathcal{G},\mathcal{D}) - {\mathcal{S}_\mathcal{A}}(\mathcal{G}',\mathcal{D})=\\
		& \log[(\prod\limits_{k = 1}^{{r_Y}} {p{{(Y|\emptyset )}^{{N_k}}}} \prod\limits_{j = 1}^{{q_T}} {\prod\limits_{k = 1}^{{r_T}} {p{{(T|X,Y)}^{{N_{jk}}}}} })  \\ 
		& /  (\prod\limits_{j = 1}^{{q_T}} {\prod\limits_{k = 1}^{{r_T}} {p{{(T|X)}^{{N_{jk}}}}} } \prod\limits_{j = 1}^{{q_Y}} {\prod\limits_{k = 1}^{{r_Y}} {p{{(Y|T)}^{{N_{jk}}}}} })]\\
		& + \Delta (\mathcal{D},\mathcal{G}') - \Delta (\mathcal{D},\mathcal{G})
	\end{aligned}
\end{equation}
Since $X \rightarrow T$ and $ T \rightarrow X$ have the same penalty term $(\Delta = {r_X}{r_T} - 1)$, and the number of log-likelihood terms in the local scores of each node is the size of $\mathcal{D}$. Writing ${p(Y|T)}$ in the above equation in the form of a posterior probability ${p(T|Y) p(Y)/p(T)}$, we can further obtain the following form:
\begin{equation}
	\begin{aligned}
		&	{\mathcal{S}_\mathcal{A}}(\mathcal{G},\mathcal{D}) - {\mathcal{S}_\mathcal{A}}(\mathcal{G}',\mathcal{D})\\ =&\log[(\prod\limits_{j = 1}^{{q_T}} {\prod\limits_{k = 1}^{{r_T}} {p{{(T|X,Y)}^{{N_{jk}}}}} } /  \prod\limits_{j = 1}^{{q_T}} {\prod\limits_{k = 1}^{{r_T}} {p{{(T|Y)}^{{N_{jk}}}}} })\\
		& / (\prod\limits_{j = 1}^{{q_T}} {\prod\limits_{k = 1}^{{r_T}} {p{{(T|X)}^{{N_{jk}}}}} } /\prod\limits_{k = 1}^{{r_T}} {p{{(T|\emptyset )}^{{N_k}}}}) ] \\
		&+ \Delta (\mathcal{D},\mathcal{G}') - \Delta (\mathcal{D},\mathcal{G})\\
		=&[{\mathcal{S}_\mathcal{A}}(X,Y \to T,\mathcal{D}) - {\mathcal{S}_\mathcal{A}}(Y \to T,\mathcal{D})] \\
		& - [{\mathcal{S}_\mathcal{A}}(X \to T,\mathcal{D}) - {\mathcal{S}_\mathcal{A}}(\emptyset  \to T,\mathcal{D})]
	\end{aligned}
\end{equation}

For the BDeu score criterion, we also use $\mathcal{G}$ and $\mathcal{G'}$ to denote local structure $X \rightarrow T \leftarrow Y$ and  $X \rightarrow T \rightarrow Y$, respectively. As above, eliminating the same local structure $\emptyset \rightarrow X$, then the BDeu score for ${\mathcal{S}_\mathcal{B}}(\mathcal{G},\mathcal{D}) - {\mathcal{S}_\mathcal{B}}(\mathcal{G'},\mathcal{D})$ is calculated as follows:
\begin{equation}
	\begin{aligned}
		&{\mathcal{S}_\mathcal{B}}(\mathcal{G},\mathcal{D}) - {\mathcal{S}_\mathcal{B}}(\mathcal{G'},\mathcal{D}) \\
		=& arg\max \frac{{p(\mathcal{D}|G)p(G)}}{{p(\mathcal{D})}} - arg\max \frac{{p(\mathcal{D}|G')p(G')}}{{p(\mathcal{D})}}\\
		=&{\mathcal{S}_\mathcal{B}}(\emptyset  \to Y,\mathcal{D}) + {\mathcal{S}_\mathcal{B}}(X,Y \to T,\mathcal{D}) \\
		&- {\mathcal{S}_\mathcal{B}}(X \to T,\mathcal{D}) - {\mathcal{S}_\mathcal{B}}(T \to Y,\mathcal{D})
	\end{aligned}
\end{equation}
We use $\mathcal{H}$ and $\mathcal{M}$ to denote the local structures $\emptyset  \to Y$ and $T  \to Y$, respectively, and $\mathcal{H'}$ and $\mathcal{M'}$ to denote the local structures $\emptyset \to T$ and $Y \to T$ respectively. Thus, for ${\mathcal{S}_\mathcal{B}}(\emptyset  \to Y,\mathcal{D}) - {\mathcal{S}_\mathcal{B}}(T \to Y,\mathcal{D})$, we can write the following form:
\begin{equation}
	\begin{aligned}
		&{\mathcal{S}_\mathcal{B}}(\emptyset  \to Y,\mathcal{D}) - {\mathcal{S}_\mathcal{B}}(T \to Y,\mathcal{D})= \\
		& \log \frac{{\Gamma (N')}}{{\Gamma (N + N')}} + \sum\limits_{k = 1}^{{r_Y}} {\log \frac{{\Gamma ({N_k} + N'/{r_Y})}}{{\Gamma (N'/{r_Y})}}}  +\log p(\mathcal{H}) \\
		&- \sum\limits_{j = 1}^{{q_Y}} {\log \frac{{\Gamma (N'/{q_Y})}}{{\Gamma ({N_j} + N'/{q_Y})}}} \\
		& - \sum\limits_{j = 1}^{{q_Y}} {\sum\limits_{k = 1}^{{r_Y}} {\log \frac{{\Gamma ({N_{jk}} + N'/{r_Y}{q_Y})}}{{\Gamma (N'/{r_Y}{q_Y})}}} }  - \log p(\mathcal{M})\\
	\end{aligned}
\end{equation}
According to Theorem 1, for nodes $T$ and $Y$, regardless of whether the true structure is $T \to Y$ or $Y \to T$, the conclusion ${\mathcal{S}_\mathcal{B}}(T \to Y, \mathcal{D}) - {\mathcal{S}_\mathcal{B}}(\emptyset \to Y, \mathcal{D}) = {\mathcal{S}_\mathcal{B}}(Y \to T, \mathcal{D}) - {\mathcal{S}_\mathcal{B}}(\emptyset \to T, \mathcal{D})$ will hold. Thus, for ${\mathcal{S}_\mathcal{B}}(\emptyset  \to Y,\mathcal{D}) - {\mathcal{S}_\mathcal{B}}(T \to Y,\mathcal{D})$, we can write the following form: 
\begin{equation}
	\begin{aligned}
		&{\mathcal{S}_\mathcal{B}}(\emptyset  \to Y,\mathcal{D}) - {\mathcal{S}_\mathcal{B}}(T \to Y,\mathcal{D})=\\
		& \log \frac{{\Gamma (N')}}{{\Gamma (N + N')}} + \sum\limits_{k = 1}^{{r_T}} {\log \frac{{\Gamma ({N_k} + N'/{r_T})}}{{\Gamma (N'/{r_T})}}} +\log p(\mathcal{H'})\\ 
		&- \sum\limits_{j = 1}^{{q_T}} {\log \frac{{\Gamma (N'/{q_T})}}{{\Gamma ({N_j} + N'/{q_T})}}}  \\
		&- \sum\limits_{j = 1}^{{q_T}} {\sum\limits_{k = 1}^{{r_T}} {\log \frac{{\Gamma ({N_{jk}} + N'/{r_T}{q_T})}}{{\Gamma (N'/{r_T}{q_T})}}} }- \log p(\mathcal{M'})\\
		=& {\mathcal{S}_\mathcal{B}}(\emptyset  \to T,\mathcal{D}) - {\mathcal{S}_\mathcal{B}}(Y \to T,\mathcal{D})
	\end{aligned}
\end{equation}
By substituting the above equation into the formula ${\mathcal{S}_\mathcal{B}}(\mathcal{G}, \mathcal{D}) - {\mathcal{S}_\mathcal{B}}(\mathcal{G'}, \mathcal{D})$, we can further represent ${\mathcal{S}_\mathcal{B}}(\mathcal{G}, \mathcal{D}) - {\mathcal{S}_\mathcal{B}}(\mathcal{G'}, \mathcal{D})$ in the following form:
\begin{equation}
	\begin{aligned}
		&{\mathcal{S}_\mathcal{B}}(\mathcal{G},\mathcal{D}) -{\mathcal{S}_\mathcal{B}}(\mathcal{G'},\mathcal{D})=\\
		&[{\mathcal{S}_\mathcal{B}}(X,Y \to T,\mathcal{D}) - {\mathcal{S}_\mathcal{B}}(Y \to T,\mathcal{D})] \\
		&- [{\mathcal{S}_\mathcal{B}}(X \to T,\mathcal{D}) - {\mathcal{S}_\mathcal{B}}(\emptyset  \to T,\mathcal{D})]
	\end{aligned}
\end{equation}
In summary, the ${\mathcal{S}_{\mathcal{A}/\mathcal{B}}}(\mathcal{G},\mathcal{D}) -{\mathcal{S}_{\mathcal{A}/\mathcal{B}}}(\mathcal{G'},\mathcal{D})$ of either the AIC score function or the BDeu score function can be written in the following form: $[{\mathcal{S}_{\mathcal{A}/\mathcal{B}}}(X,Y \to T,\mathcal{D}) - {\mathcal{S}_{\mathcal{A}/\mathcal{B}}}(Y \to T,\mathcal{D})] - [{\mathcal{S}_{\mathcal{A}/\mathcal{B}}}(X \to T,\mathcal{D}) - {\mathcal{S}_{\mathcal{A}/\mathcal{B}}}(\emptyset  \to T,\mathcal{D})]$. Where $[{\mathcal{S}_{\mathcal{A}/\mathcal{B}}}(X,Y \to T,\mathcal{D}) - {\mathcal{S}_{\mathcal{A}/\mathcal{B}}}(Y \to T,\mathcal{D})]$ can be regarded as the increase in the score of the structure \( Y \rightarrow T \) when an edge \( X \rightarrow T \) is added. Similarly, $[{\mathcal{S}_{\mathcal{A}/\mathcal{B}}}(X \to T,\mathcal{D}) - {\mathcal{S}_{\mathcal{A}/\mathcal{B}}}(\emptyset  \to T,\mathcal{D})]$ can be regarded as the increase in the score of the structure \( \emptyset \rightarrow T \) when an edge \( X \rightarrow T \) is added. However, the difference between them is that $[{\mathcal{S}_{\mathcal{A}/\mathcal{B}}}(X,Y \to T,\mathcal{D}) - {\mathcal{S}_{\mathcal{A}/\mathcal{B}}}(Y \to T,\mathcal{D})]$ has an additional edge \( Y \rightarrow T \) compared to $[{\mathcal{S}_{\mathcal{A}/\mathcal{B}}}(X \to T,\mathcal{D}) - {\mathcal{S}_{\mathcal{A}/\mathcal{B}}}(\emptyset  \to T,\mathcal{D})]$. Therefore, when  \( X \rightarrow T \leftarrow Y \) is the true structure, the $[{\mathcal{S}_{\mathcal{A}/\mathcal{B}}}(X,Y \to T,\mathcal{D}) - {\mathcal{S}_{\mathcal{A}/\mathcal{B}}}(Y \to T,\mathcal{D})]$ includes more correct parameters of the structure \( Y \rightarrow T \) than $[{\mathcal{S}_{\mathcal{A}/\mathcal{B}}}(X \to T,\mathcal{D}) - {\mathcal{S}_{\mathcal{A}/\mathcal{B}}}(\emptyset  \to T,\mathcal{D})]$. According to the score consistency, the $[{\mathcal{S}_{\mathcal{A}/\mathcal{B}}}(X,Y \to T,\mathcal{D}) - {\mathcal{S}_{\mathcal{A}/\mathcal{B}}}(Y \to T,\mathcal{D})] - [{\mathcal{S}_{\mathcal{A}/\mathcal{B}}}(X \to T,\mathcal{D}) - {\mathcal{S}_{\mathcal{A}/\mathcal{B}}}(\emptyset  \to T,\mathcal{D})] > 0$ will hold. In this case, nodes \( X, T, Y \) will be identified as a V-structure and $T$ is the collision node. Similarly, if the true structure is \( X \rightarrow T \rightarrow Y \), the $[{\mathcal{S}_{\mathcal{A}/\mathcal{B}}}(X,Y \to T,\mathcal{D}) - {\mathcal{S}_{\mathcal{A}/\mathcal{B}}}(Y \to T,\mathcal{D})]$ includes more redundant parameters of the structure \( Y \rightarrow T \) than $[{\mathcal{S}_{\mathcal{A}/\mathcal{B}}}(X \to T,\mathcal{D}) - {\mathcal{S}_{\mathcal{A}/\mathcal{B}}}(\emptyset  \to T,\mathcal{D})]$. According to the score consistency, the $[{\mathcal{S}_{\mathcal{A}/\mathcal{B}}}(X,Y \to T,\mathcal{D}) - {\mathcal{S}_{\mathcal{A}/\mathcal{B}}}(Y \to T,\mathcal{D})] - [{\mathcal{S}_{\mathcal{A}/\mathcal{B}}}(X \to T,\mathcal{D}) - {\mathcal{S}_{\mathcal{A}/\mathcal{B}}}(\emptyset  \to T,\mathcal{D})] < 0$  will hold. In this case, nodes \( X, T, Y \) will be identified as an equivalence class structure.
$\hfill\blacksquare$

With Theorem 2, we can gradually identify the final causal orientations by scoring to recognize the V-structure.

\subsection{Detailed descriptions of the HLCD algorithm}

In this section, we present the proposed HLCD algorithm through the theoretical analysis in the previous section.

\textbf{Step 1: Hybrid-based local causal skeleton construction (Lines 2-14):}  HLCD first pops a variable at the front of the queue $Q$ and assigns it to the current iteration node $Z$ (initially $Z$ is the given target $T$) (Line 4). Then, HLCD uses the constraint-based PC discovery algorithm to find the $\textbf{PC}_{Z}$ and constructs a local causal skeleton with the OR rule (Line 6). As we do not need MBs and separating sets for the edge orientation phase, HLCD can use any of the state-of-the-art PC discovery algorithms, such as MMPC, HITION-PC, FCBF, etc. Then, the HLCD stores $Z$ into \textbf{V} to prevent repeated learning of the PC of variables (line 7).  At this point, HLCD builds an initial local causal skeleton from the OR rule and learned PC sets.


As the OR rule can generate a comprehensive but redundant causal skeleton, HLCD uses the score-based method to eliminate redundant causal skeletons, ensuring they don't interfere with subsequent causal orientations. With the analysis of Theorem 1, if node $X \in \textbf{PC}_{Z}$, then the following $ \mathcal{S}(X  \rightarrow Z,\mathcal{D}) - \mathcal{S}(\phi   \rightarrow Z,\mathcal{D})  = \mathcal{S}(Z \rightarrow X,\mathcal{D}) - \mathcal{S}(\phi  \rightarrow X,\mathcal{D}) > 0 $ will be hold. HLCD does this by testing each variable $X$ in $\textbf{PC}_{Z}$ to see if it satisfies Theorem 1, and removing it from $\textbf{PC}_{Z}$ if it does not satisfy (Lines 9-13). Then, HLCD pushes all variables in $\textbf{PC}_{Z} \backslash \{ \textbf{V} \}$ into $Q$ to recursively find the PC of each node in $\textbf{PC}_{Z}$ in the next iterations for expanding (Line 14). At the end of step 1, HLCD obtains the accurate local causal skeleton consisting of all nodes in the set \textbf{V} and their PC nodes.

\begin{algorithm}[tbp]
	\footnotesize
	\caption{Hybrid Local Causal Discovery}
	
	\KwIn{$\mathcal{D}$: Data,  $T$: The target variable\;}
	\KwOut{Parents of $T$: Direct causes of $T$, Children of $T$: Direct effects of $T$\;}

	\textbf{Initialize:}  $\textbf{V} = \emptyset$, $Q$ (a regular queue) = $\{T\} $\;

	\Repeat{\rm All causal orientations of $T$ is determined, or $ Q =\emptyset $, or $\textbf{V}$ contains all variables}{

		/* Step 1: Hybrid-based local causal skeleton construction */   \
		
		$Z=Q.pop$; \
		
		\If{$Z \notin \textbf{V}$}{
			
			$\textbf{PC}_{Z}$ =  getPC($\mathcal{D}$,$Z$); \
			
			$\textbf{V}= \textbf{V} \cup \{Z\}$; \
		}

		\For{\rm each $ X \in \bf{PC_{Z}}$}{
			
			\If{\rm The local score of $X \to Z$ is not equal to $Z \to X$ \rm or $\mathcal{S}(X \to Z,\mathcal{D}) - \mathcal{S}(\phi  \to Z,\mathcal{D}) < 0 $}{
				
				$\textbf{PC}_{Z} = \textbf{PC}_{Z}\backslash \{ X \} $;  \
			}

		}

		$Q = Q.push(\textbf{PC}_{Z} \backslash \{ \textbf{V} \} )$;  \

		/* Step 2: Hybrid-based local causal orientation */    \
		
		\For{\rm each $X$,$Y \in \textbf{PC}_{Z}$}{
			
			\If{\rm The local score of $X \rightarrow Z \leftarrow Y$ is greater than $X \rightarrow Z \rightarrow Y$}{
				
				The $X$, $Y$, $Z$ form a V-structure, and $Z$ is the collision node;    \
				
			}

		}

		Using Meek-rules to orient edge orientations between variables in \textbf{V};  \
		
	}
	
	\textbf{Return} Parents of $T$, Children of $T$; \
	
\end{algorithm}

\textbf{Step 2: Hybrid-based local causal orientation (Lines 15-22):} To avoid the effect of the score equivalence, HLCD distinguishes between V-structures  and equivalence class structures by employing the score-based method. Specifically, HLCD identifies the V-structures in the causal skeleton by comparing the two local structure scores of each tuple $X$, $Y$ and $Z$ ($X, Y \in \textbf{PC}_{Z}$) in the causal skeleton obtained in step 1. If $\mathcal{S}(\phi  \rightarrow X,\mathcal{D}) + \mathcal{S}(\phi  \rightarrow Y,\mathcal{D}) + \mathcal{S}(X,Y \rightarrow Z,\mathcal{D}) > \mathcal{S}(\phi  \rightarrow X,\mathcal{D}) + \mathcal{S}(X \rightarrow Z,\mathcal{D}) + \mathcal{S}(Z \rightarrow Y,\mathcal{D})$, then the edge $X - Z$ and edge $Y - Z$ will be oriented as $X \rightarrow Z$ and $Y \rightarrow Z$ (Lines 16-20). It may be the V-structure consisting of $Z$ and its parents, or consisting of $Z$, its children, and its spouse nodes.  At this point, HLCD orients the causal orientations of all V-structures in the current causal skeleton and does not orient the causal orientations of equivalent class structures.

Finally, HLCD uses the constraint-based Meek-rule as well as the discovered V-structure to orient the causal orientations of the nodes in the set \textbf{V} (Line 21).  If all causal orientations of the $T$ are recognized in the current \textbf{V}, learning stops, otherwise it continues to expand outward  until it distinguishes between the parents and children of the $T$ (Line 22). If the set \textbf{V} includes all variables, and there are still nodes in $\textbf{PC}_T$ that have not been directed as parents or children, then these nodes are considered undirected. In this case, HLCD also outputs the undirected causal nodes. That is, if there are undirected causal orientations, HLCD outputs the local completed partially directed acyclic graph (CPDAG) of $T$.

\textbf{Theorem 3 (Correctness of HLCD)}.  Given a set of i.i.d data $\mathcal{D}$, and samples from some distribution ${\mathcal{P}}$.  As the size of the $\mathcal{D}$ goes to infinity, HLCD correctly distinguishes all parents from children of a given variable.

\textbf{\emph{Proof:}} According to Theorem 1, if $ X \in \textbf{PC}_{T} $, the local scores of $X \rightarrow T$ and $T \rightarrow X$ are the same and higher than $ \phi  \rightarrow T $ and $ \phi  \rightarrow X $. Therefore, Step 1 will keep all the true PCs found by the PC discovery algorithm and remove the false positive nodes. In the local skeleton, if $ X,Y \in \textbf{P}_{T} $, the local score of $X \rightarrow T \leftarrow Y$ is higher than that of $X \rightarrow T \rightarrow Y$ according to Theorem 2. Thus, Step 2 will find all correct V-structures in the local skeleton and will not orient the edge directions of the equivalence classes. At this point, $T$ and all its parent nodes are found correctly. Finally, the child nodes oriented out of the Meek-rule are correct. Thus, all the parents and children of a given target variable distinguished by HLCD are correct.

\subsection{Time complexity analysis of the HLCD}

Since the HLCD algorithm is recursive, we analyze its computational time complexity in both the best and worst cases. In the best-case scenario, the causal structure around the target variable can be directly identified when the target variable is a collision node. However, we must learn the causal structure in the worst-case scenario. Only then can we distinguish between the parents and children of the target variable, or, due to the existence of Markov equivalent structures, we may still be unable to differentiate between parents and children. Below, we use $|$\text{U}$|$ to denote the size of the entire variable set and $|$\text{PC}$|$ to represent the maximum size of the PC set for all nodes, respectively.
\begin{itemize}
	\item \textbf{In the best case: the target variable is the collision node.} \\
	The HLCD algorithm first uses a PC algorithm to obtain the candidate local causal skeleton for the target variable. When conditional independence tests are used, the time complexity of this step is exponential, summarized as \( O(k \cdot 2^{|\text{PC}|}) \). For example, \( k \) is equal to \( |\text{U}||\text{PC}| \) if the MMPC~\cite{tsamardinos2003time} algorithm is used, and if HITON-PC and PC-simple~\cite{aliferis2003hiton,li2015practical} are used, \( k \) is equal to \(|\text{U}|\).  When mutual information methods are used instead, the time complexity of this step is quadratic, \( O(|\text{U}||\text{PC}|) \) ~\cite{peng2005feature}.  Next, HLCD calculates the score influence of each variable in the PC set (obtained by the PC algorithm) on the target variable using Theorem 1. The time complexity of this step is \( O(|\text{PC}|) \). Finally, during skeleton orientation, HLCD pairs every two variables in the PC set and calculates their scores for forming a V-structure with the target variable as well as scores for forming an equivalence class structure. The number of such combinations is \( |\text{PC}|(|\text{PC}| - 1) / 2 \), resulting in a time complexity of \( O(|\text{PC}|^2) \). In summary, when conditional independence tests are used, the time complexity of HLCD-M (using MMPC as the PC discovery algorithm) is \( O(|\text{PC}||\text{U}|2^{|\text{PC}|} + |\text{PC}| + |\text{PC}|^2) = O(|\text{PC}||\text{U}|2^{|\text{PC}|}) \). Similarly, the time complexity of HLCD-H (using HITON-PC as the PC discovery algorithm) and HLCD-P (using PC-simple as the PC discovery algorithm) is \( O(|\text{U}|2^{|\text{PC}|}) \). When mutual information is used, the time complexity of HLCD-FS (using FCBF as the PC discovery algorithm) is \( O(|\text{PC}||\text{U}| + |\text{PC}| + |\text{PC}|^2) = O((|\text{PC}| + |\text{U}|)|\text{PC}|) \).

	\item \textbf{In the worst case: the HLCD learn the entire causal structure.} \\
	In the worst-case scenario, HLCD needs to learn the entire causal network structure to discover the local causal structure of the target variable. In this case, the algorithm must perform the above steps for each variable. Thus, the time complexity for HLCD-M in the worst case is \( O(|\text{PC}||\text{U}|^2  2^{|\text{PC}|}) \), while the time complexity for HLCD-H and HLCD-P is \( O(|\text{U}|^2  2^{|\text{PC}|}) \). For HLCD-FS, the time complexity is \( O((|\text{PC}| + |\text{U}|)|\text{PC}||\text{U}|) \).
\end{itemize}

\section{Experiments}

In this section, we conduct experiments to  evaluate the performance of our method.

\subsection{Experimental settings}

\subsubsection{Datasets}

\begin{table}[t]
	\caption{Summary of benchmark BN datasets}
	\centering
    \small
{
		\begin{tabular}{cccccc}
			\hline
			            & Num.     & Num.             &Max In/Out     & Domain     \\
			Network     & Vars     & Edges         & Degree        & Range     \\
			\hline
			Alarm       & 37       & 46                & 4/5        & 2-4      \\
			Alarm3      & 111      & 149                   & 4/5        & 2-4        \\
			Alarm5      & 185      & 265                   & 4/6        & 2-4      \\
			Alarm10     & 370      & 570                  & 4/7        & 2-4        \\
			Child       & 20       & 25                    & 2/7        & 2-6      \\
			Insurance3   & 81       & 163                 & 4/7        & 2-5       \\
			Insurance5   & 135      & 281                 & 5/8        & 2-5      \\
			Barley      & 48       & 84                   & 4/5        & 2-67       \\
			Hailfinder3    & 168      & 283                 & 5/18       & 2-11      \\
			Hailfinder5    & 280      & 458                  & 5/19       & 2-11       \\
			Hailfinder10   & 560      & 283                & 5/20       & 2-11      \\
			Link        & 724      & 1125                 & 3/14       & 2-4        \\
			Pigs        & 441      & 592                 & 2/39       & 3-3        \\
			Gene        & 801      & 972                  & 4/10       & 3-5         \\
			\hline
		\end{tabular}
	}
	\label{table.t2}
\end{table}

We use 14 benchmark BNs\footnote{These datasets are publicly available at https://pages.mtu.edu/$\sim$lebrown/supplements/ mmhc\_paper/mmhc\_index.html.} to evaluate HLCD against its rivals. Each benchmark BN contains two groups of data: one group includes 10 datasets with 500 data samples to represent small-sized dataset samples, and another group contains 10 datasets with 1000 data samples to represent medium-sized dataset samples~\cite{ling2019bamb}. A brief description of the 14 BNs is listed in Table 1. Additionally,  To evaluate the performance of these comparison algorithms in practical applications, we used a well-known dataset measuring protein and phospholipid expression in human cells~\cite{sachs2005causal} and a synthetic dataset generated by SynTReN~\cite{van2006syntren}, simulating transcriptional regulatory networks. 

\subsubsection{Comparison methods}

We compare the proposed HLCD algorithm with the following seven state-of-the-art local causal discovery algorithms, including PCD-by-PCD~\cite{yin2008partial}, MB-by-MB~\cite{wang2014discovering}, CMB~\cite{gao2015local}, LCS-FS~\cite{ling2020using},
ELCS~\cite{yang2021towards}, GraN-LCS~\cite{liang2024gradient}, and the partially causal structured discovery algorithm PSL~\cite{ling2022psl}.

\subsubsection{Evaluation metrics}

We evaluate the performance of HLCD with its competitors in terms of structural correctness, structural errors, and time efficiency.

\begin{itemize}
	\item $F1$:  ($F1=2 * Precision * Recall/(Precision+Recall)$).  Precision is the correct edges in the learning DAG divided by all edges in the learning DAG, while recall is the correct edges in the learning DAG divided by all edges in the true DAG.
	
	\item $SHD$:  Structural Hamming Distance (SHD) is the number of undirected, reversed, missing, and extra edges in the learned DAG compared to the true DAG.
	
	\item $Time$: Time is the running time (in seconds) of the local causal discovery algorithm.
\end{itemize}

It is important to note that a higher F1 score is better, while lower SHD value is preferred. In the tables,  the results are reported in the format of $A \pm B$,  where $A$ denotes the average results,  and $B$ represents the standard deviation. The best results in each setting have been marked in bold.  ``-'' means that the algorithm does not get the result in 72 hours.

\subsubsection{Implementation details}

\begin{table*}[ttbp]
	\centering
	\caption{The experimental results of F1 and SHD for HLCD and its competitors on a partial BN dataset (7 out of 14 datasets) with a sample size of 500.}
	\begin{tabular}{c|cccccccc}
		\hline
		\hline
		Metrics & Algorithm & Alarm & Child & Barley & Hailfinder3 & Link & Pigs & Gene \\
		\hline
		\multicolumn{1}{c|}{\multirow{8}[0]{*}{F1}} & GraN-LCS & 0.37$\pm$0.03 & 0.30$\pm$0.04 & 0.20$\pm$0.02 & 0.10$\pm$0.01 & -  & 0.43$\pm$0.01 & - \\
		& HLCD-M & \textbf{0.64$\pm$0.05} & \textbf{0.57$\pm$0.04} & \textbf{0.30$\pm$0.02} & \textbf{0.36$\pm$0.02} & \textbf{0.24$\pm$0.01} & \textbf{0.98$\pm$0.01} & \textbf{0.84$\pm$0.01} \\
		\cline{2-9}       & LCS-FS & 0.44$\pm$0.05 & 0.30$\pm$0.20 & 0.24$\pm$0.02 & 0.32$\pm$0.02 & 0.18$\pm$0.01 & 0.92$\pm$0.01 & 0.91$\pm$0.01 \\
		& HLCD-FS & \textbf{0.58$\pm$0.02} & \textbf{0.68$\pm$0.11} & \textbf{0.29$\pm$0.05} & \textbf{0.43$\pm$0.04} & \textbf{0.20$\pm$0.02} & \textbf{0.96$\pm$0.01} & \textbf{0.94$\pm$0.01} \\
		\cline{2-9}       & ELCS & 0.44$\pm$0.04 & 0.53$\pm$0.10 & 0.21$\pm$0.01 & 0.31$\pm$0.02 & 0.19$\pm$0.02 & 0.90$\pm$0.01 & 0.70$\pm$0.01 \\
		& HLCD-H & \textbf{0.64$\pm$0.05} & \textbf{0.66$\pm$0.19} & \textbf{0.28$\pm$0.02} & \textbf{0.37$\pm$0.02} & \textbf{0.24$\pm$0.01} & \textbf{0.98$\pm$0.00} & \textbf{0.83$\pm$0.01} \\
		\cline{2-9}       & PSL & 0.50$\pm$0.07 & 0.56$\pm$0.10 & 0.19$\pm$0.02 & 0.27$\pm$0.02 & 0.17$\pm$0.01 & 0.94$\pm$0.01 & 0.85$\pm$0.01 \\
		& HLCD-P & \textbf{0.60$\pm$0.06} & \textbf{0.70$\pm$0.05} & \textbf{0.29$\pm$0.03} & \textbf{0.34$\pm$0.03} & \textbf{0.23$\pm$0.01} & \textbf{0.99$\pm$0.00} & \textbf{0.91$\pm$0.01} \\
		\hline
		\multicolumn{1}{c|}{\multirow{8}[0]{*}{SHD}} & GraN-LCS & 2.57$\pm$0.24 & 2.41$\pm$0.16 & 5.39$\pm$0.21 & 9.40$\pm$0.44 & -  & 1.81$\pm$0.05 & - \\
		& HLCD-M & \textbf{1.29$\pm$0.16} & \textbf{1.39$\pm$0.10} & \textbf{4.93$\pm$0.18} & \textbf{4.43$\pm$0.08} & \textbf{4.05$\pm$0.09} & \textbf{0.10$\pm$0.02} & \textbf{0.48$\pm$0.03} \\
		\cline{2-9}       & LCS-FS & 1.75$\pm$0.11 & 1.85$\pm$0.53 & 4.33$\pm$0.10 & 3.02$\pm$0.06 & 4.29$\pm$0.29 & 0.47$\pm$0.08 & 0.30$\pm$0.04 \\
		& HLCD-FS & \textbf{1.43$\pm$0.11} & \textbf{0.99$\pm$0.23} & \textbf{3.05$\pm$0.15} & \textbf{2.89$\pm$0.14} & \textbf{4.09$\pm$0.25} & \textbf{0.26$\pm$0.06} & \textbf{0.25$\pm$0.05} \\
		\cline{2-9}       & ELCS & 1.76$\pm$0.14 & 1.48$\pm$0.22 & 9.31$\pm$0.44 & 5.39$\pm$0.07 & 4.53$\pm$0.09 & 0.36$\pm$0.03 & 0.83$\pm$0.02 \\
		& HLCD-H & \textbf{1.27$\pm$0.15} & \textbf{1.20$\pm$0.46} & \textbf{5.14$\pm$0.23} & \textbf{4.43$\pm$0.07} & \textbf{4.08$\pm$0.10} & \textbf{0.10$\pm$0.02} & \textbf{0.53$\pm$0.02} \\
		\cline{2-9}       & PSL & 1.62$\pm$0.18 & 1.43$\pm$0.26 & 11.52$\pm$0.69 & 5.67$\pm$0.12 & 4.41$\pm$0.08 & 0.20$\pm$0.03 & 0.41$\pm$0.03 \\
		& HLCD-P & \textbf{1.34$\pm$0.22} & \textbf{1.11$\pm$0.15} & \textbf{4.97$\pm$0.18} & \textbf{4.52$\pm$0.09} & \textbf{4.09$\pm$0.08} & \textbf{0.05$\pm$0.01} & \textbf{0.29$\pm$0.03} \\
		\hline
		\hline
	\end{tabular}%
	\label{tab:addlabel}%
\end{table*}%

All the code\footnote{The code for the causal discovery algorithm is available at https://github.com/z-dragonl/Causal-Learner.} implementations are done in Matlab~\cite{ling2022causalpackage} or Python,  and the experiments are conducted on a computer with an Intel Core i7-12700 CPU and 8GB of memory. The significance level for the CI tests is set at 0.01 and the threshold for mutual information is set at 0.03.
Furthermore, in terms of skeleton construction, existing local causal discovery algorithms such as PCD-by-PCD utilize MMPC~\cite{tsamardinos2003time}, MB-by-MB employs IAMB~\cite{tsamardinos2003algorithms}, CMB uses HITON-PC~\cite{aliferis2003hiton}, LCS-FS uses FCBF~\cite{peng2005feature}, ELCS uses HITON-PC~\cite{aliferis2003hiton}, and PSL uses PCsimple~\cite{li2015practical}. Since MMPC, HITON-PC, FCBF, and PCsimple are all PC discovery algorithms, to avoid discrepancies in results caused by the PC discovery algorithm, we maintain consistency in the PC discovery algorithm used by HLCD with its comparison algorithms and compare the experimental results separately.
In the following, we detail the four new algorithms obtained by combining HLCD with these four PC discovery algorithms.
\begin{itemize}
	\item HLCD-M: we denote HLCD combined with the PC discovery algorithm MMPC as HLCD-M and compare it experimentally with PCD-by-PCD.
	\item HLCD-FS: we denote HLCD combined with the PC discovery algorithm FCBF as HLCD-FS and compare it experimentally with LCS-FS and MB-by-MB (although IAMB, used by MB-by-MB, is not a PC discovery algorithm, it is included in the comparison here because, like FCBF, it also has linear time complexity).
	\item HLCD-H: we denote HLCD combined with the PC discovery algorithm HITON-PC as HLCD-H and compare it experimentally with CMB and ELCS.
	\item HLCD-P: we denote HLCD combined with the PC discovery algorithm PCsimple as HLCD-P and compare it experimentally with PSL.
\end{itemize}
Since GraN-LCS is different from traditional local causal discovery algorithms as it is a gradient-based method for learning local causal structure, we systematically compare it alongside HLCD-M and PCD-by-PCD.

\subsection{Experimental results}

\subsubsection{Synthetic data experiment}

\begin{figure*}[tb]
	\centering
	\includegraphics[height=1.78in]{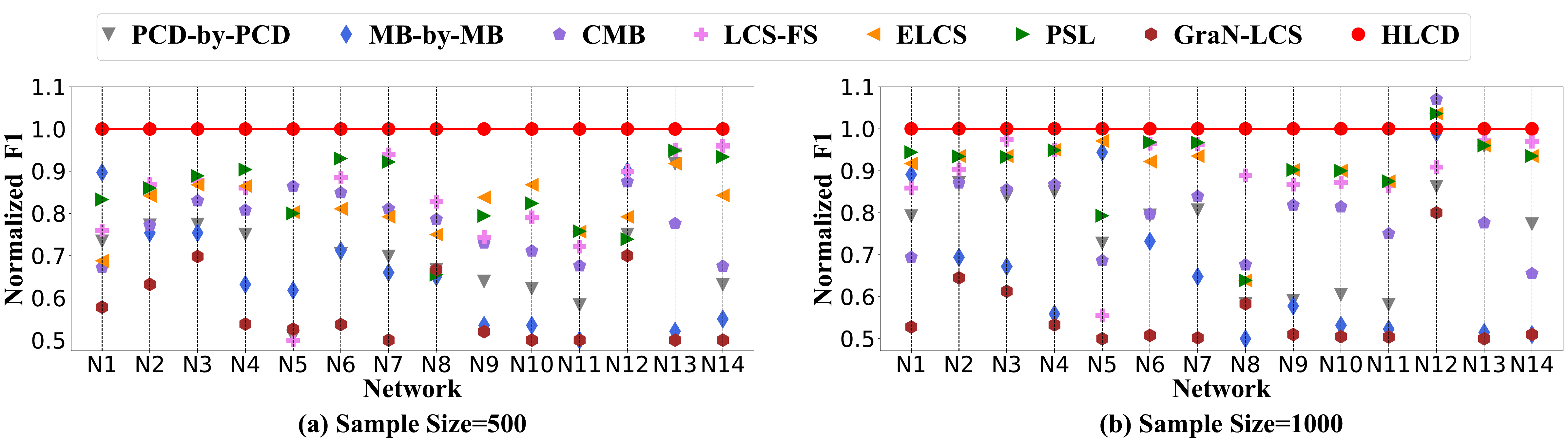}\\
	\caption{The experimental results of normalized F1, where the normalized value is the result of the comparison algorithm divided by the result of the HLCD. The larger the normalized F1, the better (the x-axis labels from N1 to N14 represent the Bayesian networks. N1: Alarm. N2: Alarm3. N3: Alarm5. N4: Alarm10. N5: Child. N6: Insurance3. N7: Insurance5. N8: Barley. N9: Hailfinder3. N10: Hailfinder5. N11: Hailfinder10. N12: Link. N13: Pigs. N14: Gene).}
\end{figure*}

\begin{figure*}[tb]
	\centering
	\includegraphics[height=1.82in]{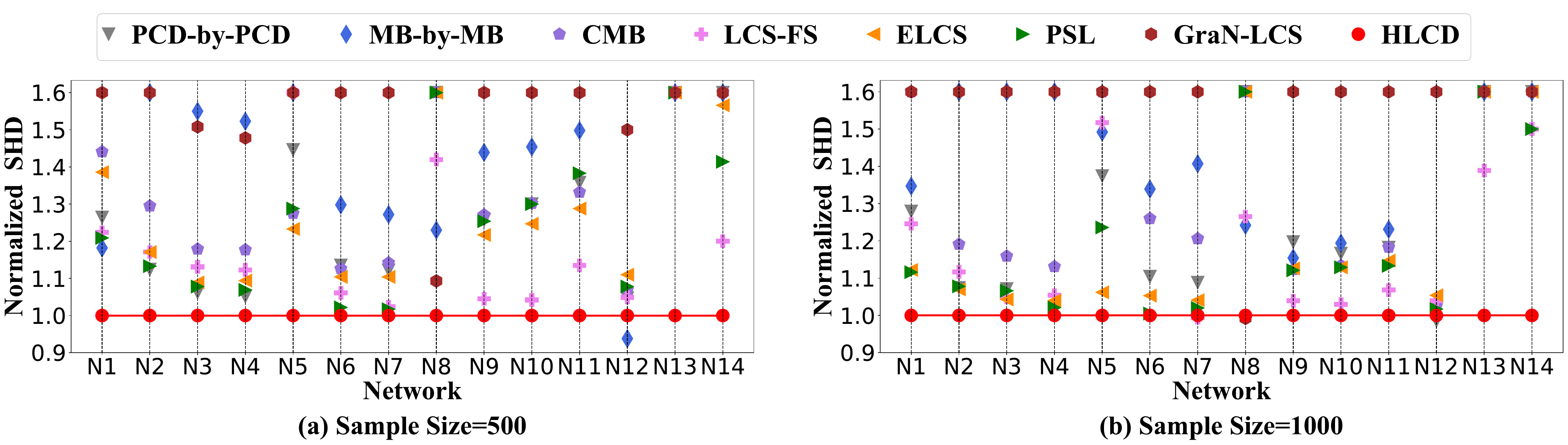}\\
	\caption{The experimental results of normalized SHD. The lower the normalized SHD, the better (the x-axis labels from N1 to N14 are identical to those in Figure 2).}
\end{figure*}

Table 2 shows the results of the F1 and SHD experiments of HLCD with state-of-the-art local causal discovery algorithms over the past four years on a partial BN dataset (7 out of 14 datasets) with a sample size of 500. The Figs. 2 and 3 present the normalized F1 and SHD results of HLCD and its competitors across 14 BN datasets. Specifically, among 28 BN sample datasets, HLCD achieves the highest F1 score in 27 datasets and the lowest SHD value in 24 datasets.  Compared to PCD-by-PCD and GraN-LCS, HLCD-M (using MMPC as the PC learning algorithm) achieves an 8\% to 27\% improvement in F1 scores and a 5\% to 20\% reduction in SHD values on networks such as Alarm, Child, Insurance, Hailfinder, and Gene. Compared to MB-by-MB and LCS-FS, HLCD-FS (using FCBF as the PC learning algorithm) achieves a 3\% to 24\% improvement in F1 scores and a 2\% to 28\% reduction in SHD values on Alarm, Insurance, Barley, Hailfinder, and Pigs networks. Compared to CMB, ELCS, and PSL, HLCD-H (using HITON-PC as the PC learning algorithm) and HLCD-P (using PCsimple as the PC learning algorithm) achieve a 4\% to 22\% improvement in F1 scores and a 6\% to 28\% reduction in SHD values on all networks.  In general, as the average number of conditional probability parameters $\Theta$ increases, the performance of local causal discovery declines due to the larger state space requiring more samples. Conversely, fewer parameters lead to better results. HLCD alleviates this issue through a scoring-based approach, achieving superior performance in both simple and complex networks.

For more detailed experimental results of HLCD on the 14 benchmark BN datasets, please refer to Tables 6-13 in the appendix following the references. 

\subsubsection{Performance evaluation of HLCD with different sample sizes}

\begin{table*}[htbp]
	\centering
	\caption{Results of F1 and SHD experiments between HLCD and its competitors on Barley's network with different sample size dimensions (Sample size: 500 $\sim$ 20000)}
	\begin{tabular}{c|ccccccc}
		\hline
		\hline
		Metrics & Algorithm & Size=500 & Size=1000 & Size=5000 & Size=10000 & Size=15000 & Size=20000 \\
		\hline
		\multirow{8}[0]{*}{F1} & GraN-LCS & 0.20$\pm$0.02 & 0.21$\pm$0.03 & 0.22$\pm$0.01 & 0.21$\pm$0.01 & 0.23$\pm$0.01 & 0.26$\pm$0.01 \\
		& HLCD-M & \textbf{0.30$\pm$0.02} & \textbf{0.36$\pm$0.02} & \textbf{0.51$\pm$0.03} & \textbf{0.52$\pm$0.01} & \textbf{0.52$\pm$0.01} & \textbf{0.55$\pm$0.00} \\
		\cline{2-8}       & LCS-FS & 0.24$\pm$0.02 & 0.32$\pm$0.02 & 0.42$\pm$0.03 & 0.44$\pm$0.02 & 0.44$\pm$0.02 & 0.44$\pm$0.01 \\
		& HLCD-FS & \textbf{0.29$\pm$0.05} & \textbf{0.36$\pm$0.02} & \textbf{0.46$\pm$0.04} & \textbf{0.49$\pm$0.01} & \textbf{0.51$\pm$0.02} & \textbf{0.50$\pm$0.01} \\
		\cline{2-8}       & ELCS & 0.21$\pm$0.01 & 0.22$\pm$0.01 & 0.39$\pm$0.02 & 0.38$\pm$0.02 & 0.38$\pm$0.01 & 0.43$\pm$0.03 \\
		& HLCD-H & \textbf{0.28$\pm$0.02} & \textbf{0.34$\pm$0.02} & \textbf{0.51$\pm$0.03} & \textbf{0.52$\pm$0.01} & \textbf{0.54$\pm$0.01} & \textbf{0.55$\pm$0.00} \\
		\cline{2-8}       & PSL & 0.19$\pm$0.02 & 0.23$\pm$0.02 & 0.41$\pm$0.02 & 0.42$\pm$0.02 & 0.42$\pm$0.02 & 0.47$\pm$0.02 \\
		& HLCD-P & \textbf{0.29$\pm$0.03} & \textbf{0.36$\pm$0.03} & \textbf{0.53$\pm$0.01} & \textbf{0.54$\pm$0.00} & \textbf{0.54$\pm$0.00} & \textbf{0.57$\pm$0.00} \\
		\hline
		\multirow{8}[0]{*}{SHD} & GraN-LCS & 5.39$\pm$0.21 & \textbf{5.22$\pm$0.23} & 4.32$\pm$0.05 & 4.43$\pm$0.17 & 4.18$\pm$0.17 & 3.62$\pm$0.14 \\
		& HLCD-M & \textbf{4.93$\pm$0.18} & 5.26$\pm$0.22 & \textbf{3.57$\pm$0.14} & \textbf{4.04$\pm$0.09} & \textbf{3.78$\pm$0.10} & \textbf{3.33$\pm$0.07} \\
		\cline{2-8}       & LCS-FS & 4.33$\pm$0.10 & 3.77$\pm$0.11 & 2.90$\pm$0.13 & 2.72$\pm$0.07 & 2.75$\pm$0.08 & 2.80$\pm$0.05 \\
		& HLCD-FS & \textbf{3.05$\pm$0.15} & \textbf{2.98$\pm$0.08} & \textbf{2.73$\pm$0.10} & \textbf{2.60$\pm$0.05} & \textbf{2.56$\pm$0.06} & \textbf{2.63$\pm$0.03} \\
		\cline{2-8}       & ELCS & 9.31$\pm$0.44 & 9.79$\pm$0.26 & 4.81$\pm$0.10 & 5.31$\pm$0.13 & 4.92$\pm$0.05 & 4.35$\pm$0.10 \\
		& HLCD-H & \textbf{5.14$\pm$0.23} & \textbf{5.28$\pm$0.21} & \textbf{3.58$\pm$0.15} & \textbf{4.05$\pm$0.11} & \textbf{3.66$\pm$0.05} & \textbf{3.30$\pm$0.08} \\
		\cline{2-8}       & PSL & 11.52$\pm$0.69 & 11.06$\pm$0.34 & 4.56$\pm$0.05 & 4.88$\pm$0.06 & 4.53$\pm$0.07 & 3.82$\pm$0.13 \\
		& HLCD-P & \textbf{4.97$\pm$0.18} & \textbf{5.23$\pm$0.24} & \textbf{3.49$\pm$0.09} & \textbf{3.89$\pm$0.07} & \textbf{3.56$\pm$0.09} & \textbf{3.09$\pm$0.04} \\
		\hline
		\hline
	\end{tabular}%
	\label{tab:addlabel}%
\end{table*}%

To evaluate the effect of sample size on the algorithm, we assessed the performance of HLCD and its competitors (state-of-the-art algorithms from the last four years) on barley networks with sample sizes ranging from 500 to 20,000. For more experimental results of HLCD under different sample sizes, please refer to Table 14 in the appendix following the references.  

Table 3 summarizes the experimental results of HLCD and its competitors on the Barley network across sample sizes ranging from 500 to 20,000. As the sample size increases, the F1 score and SHD score of all algorithms generally improve. HLCD consistently outperforms other methods across most sample sizes. Compared to GraN-LCS, HLCD-M improves the F1 score by 8–11\% and reduces the SHD by 8–10\%. Against LCS-FS, HLCD-FS increases the F1 score by 4–6\% and decreases the SHD by 4\%–18\%. When compared to ELCS, HLCD-H boosts the F1 score by 5\%–10\% and lowers the SHD by 13\%–22\%. Finally, compared to PSL, HLCD-P improves the F1 score by 10\%–12\% and reduces the SHD by about 20\%.

Further analysis reveals that methods relying on CI tests or mutual information perform poorly with smaller sample sizes but improve significantly as the sample size grows, though they still lag behind HLCD. This is because HLCD leverages score information from data to enhance performance. 

\subsubsection{Real data experiment}

\begin{table}[t]
	\centering
	\caption{Experimental results of HLCD and its competitors on Sachs dataset}
	\begin{tabular}{c|ccccc}
		\hline
		\hline
		Algorithm & F1  &\parbox[c][0.8cm][c]{1cm}{\centering Preci\\sion}   &\parbox[c][0.6cm][c]{0.6cm}{\centering Re\\call} &SHD & Time \\
		\hline
		PCD-by-PCD & 0.13  & 0.15  & 0.13  & 3.09  & \textbf{0.01 } \\
		GraN-LCS & \textbf{0.42 } & 0.45  & \textbf{0.46 } & \textbf{2.55 } & 75.30  \\
		HLCD-M & 0.37  & \textbf{0.55 } & 0.32  & \textbf{2.55 } & \textbf{0.01 } \\
		\hline
		MB-by-MB & 0.13  & 0.20  & 0.12  & 3.27  & \textbf{0.01 } \\
		LCS-FS & 0.00  & 0.00  & 0.00  & 3.09  & \textbf{0.01 } \\
		HLCD-FS & \textbf{0.37 } & \textbf{0.55 } & \textbf{0.32 } & \textbf{2.36 } & \textbf{0.01 } \\
		\hline
		CMB & 0.00  & 0.00  & 0.00  & 3.36  & \textbf{0.01 } \\
		ELCS & 0.00  & 0.00  & 0.00  & 3.45  & \textbf{0.01 } \\
		HLCD-H & \textbf{0.37 } & \textbf{0.55 } & \textbf{0.32 } & \textbf{2.55 } & \textbf{0.01 } \\
		\hline
		PSL & 0.00  & 0.00  & 0.00  & 3.45  & \textbf{0.01 } \\
		HLCD-P & \textbf{0.37 } & \textbf{0.55 } & \textbf{0.32 } & \textbf{2.55 } & \textbf{0.01 } \\
		\hline
		\hline
	\end{tabular}%
	\label{tab:addlabel}%
\end{table}%

\begin{table}[t]
	\centering
	\caption{Experimental results of HLCD and its competitors on SyNTReN dataset}
	\begin{tabular}{c|ccccc}
		\hline
		\hline
		Algorithm & F1  &\parbox[c][0.8cm][c]{1cm}{\centering Preci\\sion}   &\parbox[c][0.6cm][c]{0.6cm}{\centering Re\\call} &SHD & Time \\
		\hline
		PCD-by-PCD & \textbf{0.08 } & \textbf{0.13 } & 0.07  & \textbf{2.10 } & \textbf{0.01 } \\
		GraN-LCS & 0.01  & 0.02  & 0.01  & 3.40  & 3.46  \\
		HLCD-M & \textbf{0.08 } & 0.12  & \textbf{0.10 } & 2.40  & \textbf{0.01 } \\
		\hline
		MB-by-MB & 0.14  & 0.15  & 0.15  & 2.50  & \textbf{0.01 } \\
		LCS-FS & 0.00  & 0.00  & 0.00  & 2.90  & \textbf{0.01 } \\
		HLCD-FS & \textbf{0.25 } & \textbf{0.30 } & \textbf{0.26 } & \textbf{2.20 } & \textbf{0.01 } \\
		\hline
		CMB & 0.03  & 0.03  & 0.03  & 2.70  & \textbf{0.01 } \\
		ELCS & 0.00  & 0.00  & 0.00  & 2.75  & \textbf{0.01 } \\
		HLCD-H & \textbf{0.08 } & \textbf{0.12 } & \textbf{0.10 } & \textbf{2.40 } & \textbf{0.01 } \\
		\hline
		PSL & 0.00  & 0.00  & 0.00  & 2.75  & \textbf{0.01 } \\
		HLCD-P & \textbf{0.08 } & \textbf{0.12 } & \textbf{0.10 } & \textbf{2.40 } & \textbf{0.01 } \\

		\hline
		\hline
	\end{tabular}%
	\label{tab:addlabel}%
\end{table}%

Tables 4-5 summarize the experimental results of HLCD and its competitors on two real datasets. On the Sachs dataset, GraN-LCS achieved the highest F1 score of 42\%, followed by HLCD-M/FS/H/P at 37\%. For the SHD metric, HLCD-FS recorded the lowest value of 2.36. On the SynTReN dataset, HLCD-FS achieved the best F1 score of 25\% and the second-lowest SHD of 2.20. Furthermore, certain algorithms fail to recover correct local causal structures on the Sachs and SynTReN datasets, possibly because the learned PC or MB sets lack key collider nodes needed for accurate orientation.

\begin{figure}[tb]
	\centering
	\includegraphics[height=2.4in]{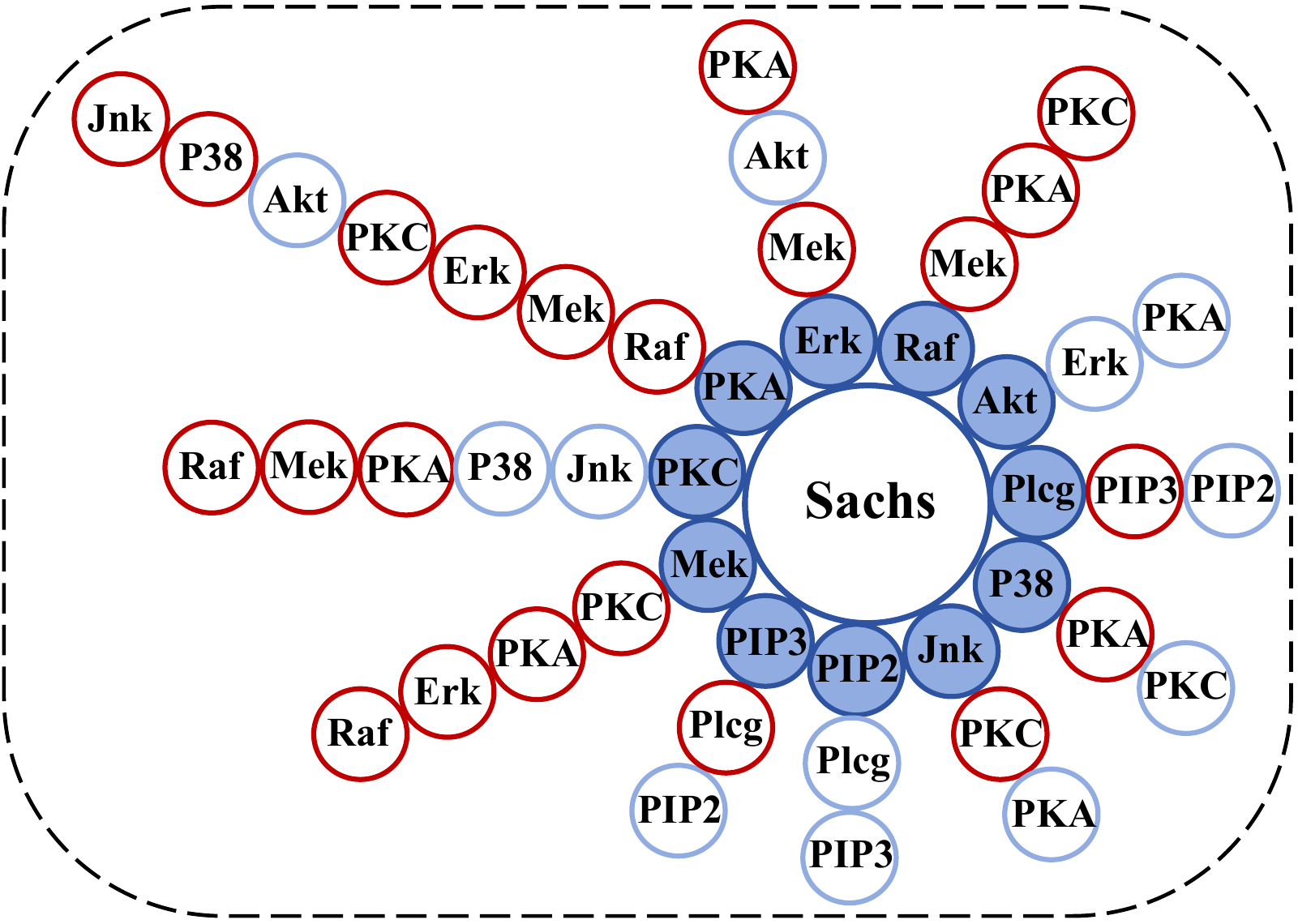}\\
	\caption{The identification results of the local causal structure for each node by the HLCD algorithm on the Sachs real network. Blue edges indicate that HLCD correctly identified a parent or child node, while red edges signify unsuccessful identification.}
\end{figure}

To illustrate the detailed learning of local causal structures by the HLCD algorithm on the Sachs network, we present the experimental results in Fig. 4. The Sachs network contains 11 nodes, shown in dark blue in the innermost circle. Each branch extending from these nodes represents their parent or child nodes. Blue edges indicate that HLCD correctly identified a parent or child node, while red edges signify unsuccessful identification. From Fig. 4, it is evident that the HLCD algorithm performs well when a node's PC set is small but struggles as the PC set size increases. This decline in performance can be attributed to two factors:  1) Insufficient recall of the  parent and child discovery algorithms: A smaller PC set may exclude correct causal nodes, causing V-structures (e.g., Raf and Mek) to be missed due to undetected parent nodes.  2) Nodes as colliders: Some nodes (e.g., PKA and PKC) are inherently collider nodes, lacking identifiable V-structures. In these cases, causal directions can only be inferred based on whether their child nodes are colliders.  

Purely CI test-based or mutual information-based methods often fail to identify local causal networks in real-world datasets accurately. GraN-LCS iteratively refines the local causal graph using an MLP but suffers from low time efficiency due to extensive matrix computations. In contrast, HLCD integrates both constraint-based and score-based methods, striking an effective balance between accuracy and efficiency.

\subsubsection{Time-efficient analysis of HLCD in synthetic data}

Tables 15-18 in the Appendix summarize the detailed time efficiency results of HLCD compared with its competitors. Specifically, HLCD-M exhibits superior time efficiency over PCD-by-PCD in most networks in a sample size of 500. Compared with GraN-LCS, HLCD-M consistently outperforms GraN-LCS in time efficiency across all networks. This is attributed to the gradient-based GraN-LCS algorithm involving extensive matrix operations, resulting in lower time efficiency. Compared to MB-by-MB and LCS-FS algorithms, HLCD-FS generally exhibits lower time efficiency than LCS-FS in most networks. HLCD-FS shows higher efficiency compared to MB-by-MB in Link, Pigs, and Gene networks, while slightly lower efficiency in Alarm and Hailfinder networks. This is because the synchronous MB learning algorithm, IAMB, has linear time complexity, giving it good time efficiency.  Compared to ELCS and CMB algorithms, HLCD-H exhibits lower time efficiency in most networks compared to ELCS, but outperforms CMB in Child, Insurance, Barley, Hailfinder, Link, and Pigs networks.  This is because ELCS reduces the number of CI tests by using the N-structure, enhancing time efficiency. Conversely, CMB uses a divide-and-conquer MB learning algorithm, which requires more time.  Against PSL, HLCD-P shows better time efficiency in high-dimensional networks but slightly weaker efficiency in low-dimensional networks.  

\subsubsection{Ablation experiments for Theorem 1 and 2}

Finally, to visually verify the validity of Theorem 1 and Theorem 2, we conducted ablation experiments on real graphs. Specifically, we applied Theorem 1 to calculate the number of non-causal nodes removed and causal nodes retained in each network. Similarly, to verify Theorem 2, we used it to determine whether all V-structures and equivalence class structures in each network were correctly identified, thus obtaining counts of correctly identified V-structures and equivalence class structures. Furthermore, to ensure the statistical significance of the experiments, we conducted these tests on networks with over 180 nodes and calculated the accuracy of each metric.

Tables 19 and 20 in the Appendix summarize the detailed ablation study results related to Theorems 1 and 2. Specifically, at a sample size of 500 and 1000, Theorem 1 removes more than 80\% to 98\% of redundant causal nodes in the Alarm, Hailfinder, Link, Pigs, and Gene networks, while retaining over 80\% to 100\% of correct causal nodes.  Regarding the identification of V-structures and equivalent class structures, at a sample size of 500 and 1000, Theorem 2 correctly identifies V-structures with accuracy rates of approximately 72\% to 100\% and equivalent class structures with accuracy rates of approximately 79\% to 100\% in the mentioned networks.

\section{Conclusion}

In this paper, we discuss the limitations of AND and OR rules in constructing exact local causal skeletons, and the problem of global causal discovery methods randomly returning incorrect local causal networks due to equivalence classes ambiguities. To address the challenges, we propose a novel hybrid local causal discovery (HLCD) method. Specifically, During the skeleton construction phase, HLCD uses maximized local scores to eliminate redundant causal skeleton structures, thereby providing a more precise causal network space. In the skeleton orientation phase, HLCD employs an innovative score-based V-structure identification approach to avoid interference caused by equivalence classes. The experimental results show that the quality of local causal discovery of HLCD is significantly better than existing methods. In future work, we aim to pursue two directions: 1) investigating the reasons behind HLCD’s superior performance in small-sample settings, and 2) extending HLCD to dynamic or time-series data analysis, as well as exploring its potential in discovering more complex causal structures.

\ifCLASSOPTIONcompsoc
  \section*{Acknowledgments}
\else
  \section*{Acknowledgment}
\fi

This work was supported by the National Key Research and Development Program of China (under grant 2021ZD0111801), the National Natural Science Foundation of China (under grant 62306002, 62272001, 62376001, 62376087, and 62120106008), the Natural Science Foundation of Anhui Province of China under Grant 2108085QF270, and the Xunfei Zhiyuan Digital Transformation Innovation Research Special for Universities (2023ZY001).

\ifCLASSOPTIONcaptionsoff
  \newpage
\fi

\bibliographystyle{IEEEtran}
\bibliography{references}

\clearpage
\appendix

\begin{table*}[b]
	\centering
	\caption{Summary of structural correctness and errors experiment data for HLCD-M and its competitors PCD-by-PCD and GraN-LCS (Sample size: 500)}
	\renewcommand{\arraystretch}{1.14}
	\resizebox{172mm}{!}{
		
\begin{tabular}{cc||ccc|ccccc}
			\hline
			\hline
			Network & Algorithm & F1 (\textuparrow) & Precision (\textuparrow) & Recall (\textuparrow) & SHD (\textdownarrow) & Undirected (\textdownarrow) & Reverse (\textdownarrow) & Miss (\textdownarrow) & Extra (\textdownarrow) \\
			\hline
			\multirow{3}[2]{*}{Alarm} & PCD-by-PCD & 0.47$\pm$0.04 & 0.51$\pm$0.05 & 0.46$\pm$0.04 & 1.63$\pm$0.16 & 0.27$\pm$0.09 & 0.30$\pm$0.09 & 0.72$\pm$0.09 & 0.34$\pm$0.09 \\
			& GraN-LCS & 0.37$\pm$0.03 & 0.38$\pm$0.04 & 0.40$\pm$0.03 & 2.57$\pm$0.24 & 0.00$\pm$0.00 & 0.58$\pm$0.08 & 0.98$\pm$0.08 & 1.01$\pm$0.21 \\
			& HLCD-M & \textbf{0.64$\pm$0.05} & 0.67$\pm$0.05 & 0.65$\pm$0.05 & \textbf{1.29$\pm$0.16} & 0.20$\pm$0.07 & 0.13$\pm$0.09 & 0.51$\pm$0.06 & 0.45$\pm$0.07 \\
			\hline
			\multirow{3}[2]{*}{Alarm3} & PCD-by-PCD & 0.44$\pm$0.03 & 0.50$\pm$0.03 & 0.41$\pm$0.03 & 1.72$\pm$0.07 & 0.34$\pm$0.07 &0.33$\pm$0.03 & 0.87$\pm$0.04 & 0.18$\pm$0.04\\
			& GraN-LCS & 0.36$\pm$0.01 & 0.38$\pm$0.01 & 0.37$\pm$0.01 & 2.63$\pm$0.02 & 0.00$\pm$0.00 & 0.51$\pm$0.03 & 1.18$\pm$0.02 & 0.94$\pm$0.02 \\
			& HLCD-M & \textbf{0.57$\pm$0.03} & 0.61$\pm$0.04 & 0.56$\pm$0.03 & \textbf{1.53$\pm$0.12} & 0.24$\pm$0.09 & 0.21$\pm$0.03 & 0.67$\pm$0.04 & 0.41$\pm$0.06 \\
			\hline
			\multirow{3}[2]{*}{Alarm5} & PCD-by-PCD & 0.41$\pm$0.01 & 0.48$\pm$0.02 & 0.39$\pm$0.01 & 1.94$\pm$0.05 & 0.38$\pm$0.04 & 0.35$\pm$0.04 & 1.03$\pm$0.04 & 0.18$\pm$0.02\\
\hline
			\hline
		\end{tabular}%
	}
	\label{tab:1}%
\end{table*}%

\begin{table*}[htb]
	\centering
	\caption{Summary of structural correctness and errors experiment data for HLCD-M and its competitors PCD-by-PCD and GraN-LCS (Sample size: 1000)}
	\renewcommand{\arraystretch}{1.16}
	\resizebox{172mm}{!}{
		
		\begin{tabular}{cc||ccc|ccccc}
			\hline
			\hline
			Network & Algorithm & F1 (\textuparrow) & Precision (\textuparrow) & Recall (\textuparrow) & SHD (\textdownarrow) & Undirected (\textdownarrow) & Reverse (\textdownarrow) & Miss (\textdownarrow) & Extra (\textdownarrow) \\
			\hline
			\multirow{3}[2]{*}{Alarm} & PCD-by-PCD & 0.57$\pm$0.07 & 0.63$\pm$0.07 & 0.54$\pm$0.07 & 1.10$\pm$0.14 & 0.25$\pm$0.14 & 0.32$\pm$0.10 & 0.51$\pm$0.08 & 0.02$\pm$0.03 \\
			& GraN-LCS & 0.38$\pm$0.03 & 0.38$\pm$0.03 & 0.42$\pm$0.03 & 2.43$\pm$0.14 & 0.00$\pm$0.00 & 0.58$\pm$0.07 & 0.94$\pm$0.08 & 0.91$\pm$0.10 \\
			& HLCD-M & \textbf{0.72$\pm$0.05} & 0.74$\pm$0.05 & 0.71$\pm$0.06 & \textbf{0.86$\pm$0.16} & 0.16$\pm$0.08 & 0.15$\pm$0.08 & 0.35$\pm$0.07 & 0.20$\pm$0.05 \\
			\hline
			\multirow{3}[2]{*}{Alarm3} & PCD-by-PCD & 0.54$\pm$0.04 & 0.60$\pm$0.04 & 0.50$\pm$0.04 & 1.30$\pm$0.08 & 0.27$\pm$0.08 & 0.38$\pm$0.04 & 0.65$\pm$0.03 & 0.01$\pm$0.01 \\
			& GraN-LCS & 0.40$\pm$0.01 & 0.44$\pm$0.02 & 0.40$\pm$0.01 & 2.35$\pm$0.08 & 0.00$\pm$0.00 & 0.52$\pm$0.04 & 1.13$\pm$0.03 & 0.70$\pm$0.07 \\
			& HLCD-M & \textbf{0.62$\pm$0.03} & 0.66$\pm$0.03 & 0.60$\pm$0.03 & \textbf{1.21$\pm$0.09} & 0.30$\pm$0.11 & 0.20$\pm$0.06 & 0.52$\pm$0.03 & 0.18$\pm$0.03 \\
			\hline
			\multirow{3}[2]{*}{Alarm5} & PCD-by-PCD & 0.52$\pm$0.04 & 0.58$\pm$0.04 & 0.48$\pm$0.04 & 1.51$\pm$0.09 & 0.28$\pm$0.06 & 0.40$\pm$0.03 & 0.82$\pm$0.05 & 0.01$\pm$0.01 \\
\hline
			\hline
		\end{tabular}%
	}
	\label{tab:2}%
\end{table*}%

\begin{table*}[htb]
	\centering
	\caption{Summary of structural correctness and errors experiment data for HLCD-FS and its competitors MB-by-MB and LCS-FS (Sample size: 500)}
	\renewcommand{\arraystretch}{1.16}
	\resizebox{172mm}{!}{
		
		\begin{tabular}{cc||ccc|ccccc}
			\hline
			\hline
			Network & Algorithm & F1 (\textuparrow) & Precision (\textuparrow) & Recall (\textuparrow) & SHD (\textdownarrow) & Undirected (\textdownarrow) & Reverse (\textdownarrow) & Miss (\textdownarrow) & Extra (\textdownarrow)  \\
			\hline
			
			\multirow{3}[2]{*}{Alarm} & MB-by-MB & 0.52$\pm$0.06 & 0.55$\pm$0.05 & 0.52$\pm$0.06 & 1.69$\pm$0.12 & 0.07$\pm$0.08 & 0.46$\pm$0.09 & 0.73$\pm$0.07 & 0.43$\pm$0.07  \\
			& LCS-FS & 0.44$\pm$0.05 & 0.48$\pm$0.05 & 0.43$\pm$0.05 & 1.75$\pm$0.11 & 0.33$\pm$0.11 & 0.25$\pm$0.08 & 0.80$\pm$0.05 & 0.38$\pm$0.08  \\
			& HLCD-FS & \textbf{0.58$\pm$0.02} & 0.63$\pm$0.02 & 0.57$\pm$0.03 & \textbf{1.43$\pm$0.11} & 0.09$\pm$0.04 & 0.14$\pm$0.05 & 0.80$\pm$0.05 & 0.40$\pm$0.07  \\
			\hline
			\multirow{3}[2]{*}{Alarm3} & MB-by-MB & 0.46$\pm$0.03 & 0.50$\pm$0.03 & 0.45$\pm$0.03 & 2.07$\pm$0.09 & 0.06$\pm$0.05 & 0.50$\pm$0.07 & 0.99$\pm$0.04 & 0.52$\pm$0.06  \\
			& LCS-FS & 0.53$\pm$0.02 & 0.61$\pm$0.03 & 0.50$\pm$0.02 & 1.51$\pm$0.06 & 0.23$\pm$0.06 & 0.32$\pm$0.08 & 0.78$\pm$0.03 & 0.20$\pm$0.05  \\
			& HLCD-FS & \textbf{0.61$\pm$0.03} & 0.70$\pm$0.04 & 0.57$\pm$0.03 & \textbf{1.29$\pm$0.08} & 0.20$\pm$0.12 & 0.12$\pm$0.07 & 0.77$\pm$0.03 & 0.19$\pm$0.05  \\
			\hline
			\multirow{3}[2]{*}{Alarm5} & MB-by-MB & 0.43$\pm$0.03 & 0.48$\pm$0.03 & 0.42$\pm$0.03 & 2.48$\pm$0.10 & 0.03$\pm$0.02 & 0.56$\pm$0.04 & 1.18$\pm$0.03 & 0.70$\pm$0.06 \\
\hline
			\hline
			
		\end{tabular}%
	}
	\label{tab:3}%
\end{table*}%

\begin{table*}[htb]
	\centering
	\caption{Summary of structural correctness and errors experiment data for HLCD-FS and its competitors MB-by-MB and LCS-FS (Sample size: 1000)}
	\renewcommand{\arraystretch}{1.16}
	\resizebox{172mm}{!}{
		
		\begin{tabular}{cc||ccc|ccccc}
			\hline
			\hline
			Network & Algorithm & F1 (\textuparrow) & Precision (\textuparrow) & Recall (\textuparrow) & SHD (\textdownarrow) & Undirected (\textdownarrow) & Reverse (\textdownarrow) & Miss (\textdownarrow) & Extra (\textdownarrow) \\
			\hline
			
			\multirow{3}[2]{*}{Alarm} & MB-by-MB & 0.57$\pm$0.04 & 0.58$\pm$0.05 & 0.60$\pm$0.03 & 1.59$\pm$0.17 & 0.20$\pm$0.16 & 0.40$\pm$0.09 & 0.49$\pm$0.05 & 0.50$\pm$0.14 \\
			& LCS-FS  & 0.55$\pm$0.04 & 0.59$\pm$0.04 & 0.54$\pm$0.04 & 1.47$\pm$0.13 & 0.22$\pm$0.11 & 0.29$\pm$0.11 & 0.63$\pm$0.04 & 0.34$\pm$0.06 \\
			& HLCD-FS  & \textbf{0.64$\pm$0.03} & 0.68$\pm$0.04 & 0.63$\pm$0.03 & \textbf{1.18$\pm$0.12} & 0.14$\pm$0.04 & 0.06$\pm$0.05 & 0.63$\pm$0.05 & 0.34$\pm$0.07 \\
			\hline
			\multirow{3}[2]{*}{Alarm3} & MB-by-MB  & 0.43$\pm$0.01 & 0.43$\pm$0.02 & 0.47$\pm$0.01 & 2.44$\pm$0.10 & 0.04$\pm$0.04 & 0.71$\pm$0.07 & 0.72$\pm$0.02 & 0.97$\pm$0.08 \\
			& LCS-FS  & 0.56$\pm$0.02 & 0.63$\pm$0.02 & 0.53$\pm$0.03 & 1.34$\pm$0.08 & 0.31$\pm$0.06 & 0.22$\pm$0.03 & 0.66$\pm$0.02 & 0.15$\pm$0.04 \\
			& HLCD-FS & \textbf{0.62$\pm$0.03} & 0.69$\pm$0.03 & 0.58$\pm$0.03 & \textbf{1.20$\pm$0.08} & 0.27$\pm$0.11 & 0.12$\pm$0.04 & 0.65$\pm$0.03 & 0.15$\pm$0.04 \\
			\hline
			\multirow{3}[2]{*}{Alarm5} & MB-by-MB  & 0.39$\pm$0.02 & 0.39$\pm$0.02 & 0.43$\pm$0.02 & 2.91$\pm$0.08 & 0.02$\pm$0.01 & 0.79$\pm$0.04 & 0.94$\pm$0.02 & 1.17$\pm$0.07 \\
\hline
			\hline
			
		\end{tabular}%
	}
	\label{tab:4}%
\end{table*}%

\begin{table*}[htb]
	\centering
	\caption{Summary of structural correctness and errors experiment data for HLCD-H and its competitors CMB and ELCS (Sample size: 500)}
	\renewcommand{\arraystretch}{1.16}
	\resizebox{172mm}{!}{
		
		\begin{tabular}{cc||ccc|ccccc}
			\hline
			\hline
			Network & Algorithm & F1 (\textuparrow) & Precision (\textuparrow) & Recall (\textuparrow) & SHD (\textdownarrow) & Undirected (\textdownarrow) & Reverse (\textdownarrow) & Miss (\textdownarrow) & Extra (\textdownarrow) \\
			\hline
			
			\multirow{3}[2]{*}{Alarm} & CMB & 0.43$\pm$0.09 & 0.44$\pm$0.09 & 0.43$\pm$0.09 & 1.83$\pm$0.23 & 0.25$\pm$0.23 & 0.65$\pm$0.06 & 0.49$\pm$0.09 & 0.44$\pm$0.08 \\
			& ELCS & 0.44$\pm$0.04 & 0.47$\pm$0.05 & 0.43$\pm$0.04 & 1.76$\pm$0.14 & 0.46$\pm$0.08 & 0.36$\pm$0.04 & 0.52$\pm$0.05 & 0.42$\pm$0.08 \\
			& HLCD-H & \textbf{0.64$\pm$0.05} & 0.66$\pm$0.05 & 0.65$\pm$0.05 & \textbf{1.27$\pm$0.15} & 0.21$\pm$0.08 & 0.15$\pm$0.10 & 0.49$\pm$0.06 & 0.43$\pm$0.08 \\
			\hline
			\multirow{3}[2]{*}{Alarm3} & CMB & 0.44$\pm$0.03 & 0.47$\pm$0.03 & 0.44$\pm$0.03 & 1.89$\pm$0.13 & 0.21$\pm$0.07 & 0.63$\pm$0.13 & 0.68$\pm$0.03 & 0.37$\pm$0.07 \\
			& ELCS & 0.48$\pm$0.03 & 0.52$\pm$0.03 & 0.46$\pm$0.02 & 1.71$\pm$0.10 & 0.38$\pm$0.10 & 0.31$\pm$0.07 & 0.67$\pm$0.02 & 0.35$\pm$0.06 \\
			& HLCD-H & \textbf{0.57$\pm$0.03} & 0.61$\pm$0.03 & 0.56$\pm$0.03 & \textbf{1.46$\pm$0.08} & 0.23$\pm$0.07 & 0.23$\pm$0.04 & 0.64$\pm$0.02 & 0.36$\pm$0.07 \\
			\hline
			\multirow{3}[2]{*}{Alarm5} & CMB & 0.44$\pm$0.02 & 0.47$\pm$0.02 & 0.43$\pm$0.03 & 2.11$\pm$0.07 & 0.20$\pm$0.06 & 0.65$\pm$0.06 & 0.86$\pm$0.03 & 0.40$\pm$0.04 \\
\hline
			\hline
			
		\end{tabular}%
	}
	\label{tab:5}%
\end{table*}%

\begin{table*}[htb]
	\centering
	\caption{Summary of structural correctness and errors experiment data for HLCD-H and its competitors CMB and ELCS (Sample size: 1000)}
	\renewcommand{\arraystretch}{1.16}
	\resizebox{172mm}{!}{
		
		\begin{tabular}{cc||ccc|ccccc}
			\hline
			\hline
			Network & Algorithm & F1 (\textuparrow) & Precision (\textuparrow) & Recall (\textuparrow) & SHD (\textdownarrow) & Undirected (\textdownarrow) & Reverse (\textdownarrow) & Miss (\textdownarrow) & Extra (\textdownarrow) \\
			\hline
			
			\multirow{3}[2]{*}{Alarm} & CMB  & 0.50$\pm$0.07 & 0.52$\pm$0.07 & 0.50$\pm$0.08 & 1.44$\pm$0.19 & 0.16$\pm$0.15 & 0.75$\pm$0.10 & 0.34$\pm$0.07 & 0.20$\pm$0.07 \\
			& ELCS  & 0.66$\pm$0.05 & 0.69$\pm$0.05 & 0.65$\pm$0.06 & 0.92$\pm$0.16 & 0.34$\pm$0.09 & 0.08$\pm$0.04 & 0.34$\pm$0.07 & 0.16$\pm$0.05 \\
			& HLCD-H  & \textbf{0.72$\pm$0.05} & 0.74$\pm$0.05 & 0.71$\pm$0.06 & \textbf{0.82$\pm$0.16} & 0.16$\pm$0.08 & 0.15$\pm$0.07 & 0.33$\pm$0.07 & 0.17$\pm$0.06 \\
			\hline
			\multirow{3}[2]{*}{Alarm3} & CMB & 0.54$\pm$0.03 & 0.57$\pm$0.03 & 0.53$\pm$0.03 & 1.37$\pm$0.11 & 0.14$\pm$0.06 & 0.58$\pm$0.08 & 0.51$\pm$0.02 & 0.14$\pm$0.03 \\
			& ELCS  & 0.58$\pm$0.03 & 0.63$\pm$0.04 & 0.56$\pm$0.03 & 1.23$\pm$0.08 & 0.45$\pm$0.12 & 0.14$\pm$0.07 & 0.51$\pm$0.02 & 0.14$\pm$0.03 \\
			& HLCD-H  & \textbf{0.62$\pm$0.03} & 0.67$\pm$0.04 & 0.60$\pm$0.03 & \textbf{1.15$\pm$0.10} & 0.32$\pm$0.10 & 0.20$\pm$0.06 & 0.49$\pm$0.02 & 0.15$\pm$0.03 \\
			\hline
			\multirow{3}[2]{*}{Alarm5} & CMB  & 0.53$\pm$0.02 & 0.57$\pm$0.03 & 0.52$\pm$0.02 & 1.60$\pm$0.08 & 0.11$\pm$0.03 & 0.62$\pm$0.06 & 0.69$\pm$0.02 & 0.18$\pm$0.03 \\
\hline
			\hline
			
		\end{tabular}%
	}
	\label{tab:6}%
\end{table*}%

\begin{table*}[htb]
	\centering
	\caption{Summary of structural correctness and errors experiment data for HLCD-P and its competitor PSL (Sample size: 500)}
	\resizebox{172mm}{!}{

		\begin{tabular}{cc||ccc|ccccc}
			\hline
			\hline
			Network & Algorithm & F1 (\textuparrow) & Precision (\textuparrow) & Recall (\textuparrow) & SHD (\textdownarrow) & Undirected (\textdownarrow) & Reverse (\textdownarrow) & Miss (\textdownarrow) & Extra (\textdownarrow) \\
			\hline
			\multirow{2}[2]{*}{Alarm} & PSL & 0.50$\pm$0.07 & 0.53$\pm$0.08 & 0.50$\pm$0.07 & 1.62$\pm$0.18 & 0.32$\pm$0.15 & 0.34$\pm$0.09 & 0.53$\pm$0.09 & 0.44$\pm$0.09 \\
			& HLCD-P & \textbf{0.60$\pm$0.06} & 0.63$\pm$0.05 & 0.61$\pm$0.06 & \textbf{1.34$\pm$0.22} & 0.24$\pm$0.09 & 0.14$\pm$0.06 & 0.54$\pm$0.08 & 0.43$\pm$0.09 \\
			\hline
			\multirow{2}[2]{*}{Alarm3} & PSL & 0.49$\pm$0.03 & 0.53$\pm$0.03 & 0.47$\pm$0.03 & 1.62$\pm$0.11 & 0.28$\pm$0.06 & 0.43$\pm$0.07 & 0.63$\pm$0.03 & 0.29$\pm$0.05 \\
			& HLCD-P & \textbf{0.57$\pm$0.03} & 0.62$\pm$0.03 & 0.56$\pm$0.03 & \textbf{1.43$\pm$0.06} & 0.31$\pm$0.09 & 0.20$\pm$0.04 & 0.62$\pm$0.03 & 0.29$\pm$0.05 \\
			\hline
			\multirow{2}[2]{*}{Alarm5} & PSL & 0.48$\pm$0.03 & 0.53$\pm$0.03 & 0.47$\pm$0.03 & 1.83$\pm$0.06 & 0.31$\pm$0.09 & 0.38$\pm$0.05 & 0.82$\pm$0.03 & 0.32$\pm$0.04 \\
			& HLCD-P & \textbf{0.54$\pm$0.02} & 0.59$\pm$0.02 & 0.52$\pm$0.02 & \textbf{1.70$\pm$0.05} & 0.30$\pm$0.08 & 0.26$\pm$0.03 & 0.81$\pm$0.02 & 0.33$\pm$0.04 \\
			\hline
			\multirow{2}[2]{*}{Alarm10} & PSL & 0.47$\pm$0.01 & 0.54$\pm$0.01 & 0.45$\pm$0.01 & 2.07$\pm$0.04 & 0.26$\pm$0.04 & 0.40$\pm$0.03 & 1.02$\pm$0.02 & 0.38$\pm$0.02 \\
\hline
			\hline
		\end{tabular}%
		
	}
	\label{tab:7}%
\end{table*}%

\begin{table*}[htb]
	\centering
	\caption{Summary of structural correctness and errors experiment data for HLCD-P and its competitor PSL (Sample size: 1000)}
	\resizebox{172mm}{!}{

		\begin{tabular}{cc||ccc|ccccc}
			\hline
			\hline
			Network & Algorithm & F1 (\textuparrow) & Precision (\textuparrow) & Recall (\textuparrow) & SHD (\textdownarrow) & Undirected (\textdownarrow) & Reverse (\textdownarrow) & Miss (\textdownarrow) & Extra (\textdownarrow) \\
			\hline
			\multirow{2}[2]{*}{Alarm} & PSL  & 0.67$\pm$0.06 & 0.70$\pm$0.05 & 0.65$\pm$0.06 & 0.96$\pm$0.19 & 0.21$\pm$0.11 & 0.19$\pm$0.07 & 0.36$\pm$0.08 & 0.20$\pm$0.08 \\
			& HLCD-P & \textbf{0.71$\pm$0.04} & 0.73$\pm$0.03 & 0.70$\pm$0.03 & \textbf{0.86$\pm$0.14} & 0.14$\pm$0.03 & 0.16$\pm$0.07 & 0.37$\pm$0.08 & 0.19$\pm$0.08 \\
			\hline
			\multirow{2}[2]{*}{Alarm3} & PSL & 0.57$\pm$0.03 & 0.61$\pm$0.03 & 0.54$\pm$0.03 & 1.25$\pm$0.08 & 0.38$\pm$0.15 & 0.26$\pm$0.09 & 0.50$\pm$0.03 & 0.10$\pm$0.03 \\
			& HLCD-P & \textbf{0.61$\pm$0.02} & 0.66$\pm$0.03 & 0.59$\pm$0.02 & \textbf{1.16$\pm$0.08} & 0.38$\pm$0.09 & 0.18$\pm$0.06 & 0.50$\pm$0.03 & 0.10$\pm$0.03 \\
			\hline
			\multirow{2}[2]{*}{Alarm5} & PSL & 0.56$\pm$0.01 & 0.62$\pm$0.02 & 0.54$\pm$0.02 & 1.46$\pm$0.06 & 0.36$\pm$0.07 & 0.27$\pm$0.05 & 0.69$\pm$0.02 & 0.14$\pm$0.02 \\
			& HLCD-P & \textbf{0.60$\pm$0.02} & 0.66$\pm$0.02 & 0.58$\pm$0.02 & \textbf{1.37$\pm$0.07} & 0.31$\pm$0.09 & 0.24$\pm$0.05 & 0.69$\pm$0.02 & 0.14$\pm$0.03 \\
			\hline
			\multirow{2}[2]{*}{Alarm10} & PSL  & 0.56$\pm$0.02 & 0.64$\pm$0.02 & 0.53$\pm$0.01 & 1.62$\pm$0.05 & 0.29$\pm$0.07 & 0.31$\pm$0.06 & 0.85$\pm$0.02 & 0.17$\pm$0.01 \\
\hline
			\hline
		\end{tabular}%
		
	}
	\label{tab:8}%
\end{table*}%

\begin{table*}[t]
	\centering
	\caption{Results of F1 and SHD experiments between HLCD and its competitors on Barley's network with different sample size dimensions (Sample size: 500 $\sim$ 20000)}
	\renewcommand{\arraystretch}{1.15}
	\resizebox{162mm}{!}{
		\begin{tabular}{c|ccccccc}
			\hline
			\hline
			Metrics & Algorithm & Size=500 & Size=1000 & Size=5000 & Size=10000 & Size=15000 & Size=20000 \\
			\hline
			\multirow{11}[0]{*}{F1} & PCD-by-PCD & 0.20$\pm$0.02 & 0.21$\pm$0.01 & 0.42$\pm$0.02 & 0.42$\pm$0.03 & 0.45$\pm$0.02 & 0.49$\pm$0.01 \\
			& GraN-LCS & 0.20$\pm$0.02 & 0.21$\pm$0.03 & 0.22$\pm$0.01 & 0.21$\pm$0.01 & 0.23$\pm$0.01 & 0.26$\pm$0.01 \\
			& HLCD-M & \textbf{0.30$\pm$0.02} & \textbf{0.36$\pm$0.02} & \textbf{0.51$\pm$0.03} & \textbf{0.52$\pm$0.01} & \textbf{0.52$\pm$0.01} & \textbf{0.55$\pm$0.00} \\
			\cline{2-8}       & MB-by-MB & 0.00$\pm$0.00 & 0.07$\pm$0.02 & 0.35$\pm$0.03 & 0.47$\pm$0.05 & 0.47$\pm$0.06 & \textbf{0.52$\pm$0.02} \\
			& LCS-FS & 0.24$\pm$0.02 & 0.32$\pm$0.02 & 0.42$\pm$0.03 & 0.44$\pm$0.02 & 0.44$\pm$0.02 & 0.44$\pm$0.01 \\
			& HLCD-FS & \textbf{0.29$\pm$0.05} & \textbf{0.36$\pm$0.02} & \textbf{0.46$\pm$0.04} & \textbf{0.49$\pm$0.01} & \textbf{0.51$\pm$0.02} & 0.51$\pm$0.01 \\
			\cline{2-8}       & CMB & 0.22$\pm$0.03 & 0.23$\pm$0.01 & 0.40$\pm$0.06 & 0.42$\pm$0.02 & 0.45$\pm$0.02 & 0.51$\pm$0.01 \\
\hline
			\hline
		\end{tabular}%
		
	}
	\label{tab:addlabel}%
\end{table*}%

\begin{table*}[htb]
	\centering
	\footnotesize
	\caption{Summary of runtimes (in seconds) for HLCD-M and its rivals PCD-by-PCD and GraN-LCS with sample sizes of 500 and 1000}
	{
		\begin{tabular}{cc|ccc|ccc}
			\hline
			\hline
			&& Size=500 & & &  Size=1000	&   \\
			& Network & PCD-by-PCD & GraN-LCS & HLCD-M & PCD-by-PCD & GraN-LCS & HLCD-M \\
			
			\hline
			& Alarm & 0.02$\pm$0.00 & 9.94$\pm$0.92 & \textbf{0.01$\pm$0.00} & 0.02$\pm$0.00 & 13.00$\pm$1.32 & \textbf{0.01$\pm$0.00}\\
			& Alarm3 & 0.07$\pm$0.02 & 13.18$\pm$0.05 & \textbf{0.06$\pm$0.02} & \textbf{0.08$\pm$0.02} & 20.65$\pm$1.00 & 0.09$\pm$0.02 \\
			& Alarm5  & 0.22$\pm$0.05 & 19.23$\pm$3.28 & \textbf{0.14$\pm$0.03} & \textbf{0.16$\pm$0.03} & 27.77$\pm$2.21 & \textbf{0.16$\pm$0.05} \\
			& Alarm10 & 1.37$\pm$0.51 & 81.70$\pm$1.25 & \textbf{0.46$\pm$0.14} & \textbf{0.68$\pm$0.10} & 94.63$\pm$2.05 & 0.80$\pm$0.22 \\
			& Child & \textbf{0.01$\pm$0.00} & 13.42$\pm$1.97 & 0.02$\pm$0.00 & \textbf{0.02$\pm$0.00} & 13.82$\pm$1.45 & 0.03$\pm$0.00 \\
			& Insurance3 & \textbf{0.03$\pm$0.01} & 19.45$\pm$1.19 & 0.07$\pm$0.01 & \textbf{0.04$\pm$0.01} & 21.08$\pm$0.99 & 0.09$\pm$0.02 \\
\hline
			\hline
		\end{tabular}%

	}
	\label{tab:9}%
	
\end{table*}%

\begin{table*}[htb]
	\centering
	\footnotesize
	\caption{Summary of runtime (in seconds) for HLCD-FS and its rivals MB-by-MB and LCS-FS with sample sizes of 500 and 1000}
	{
		\begin{tabular}{cc|ccc|ccc}
			\hline
			\hline
			&&   Size=500 & & &  Size=1000	&\\
			& Network & MB-by-MB & LCS-FS & HLCD-FS & MB-by-MB & LCS-FS & HLCD-FS \\
			
			\hline
			& Alarm & \textbf{0.01$\pm$0.00} & \textbf{0.01$\pm$0.00} & \textbf{0.01$\pm$0.00} & \textbf{0.01$\pm$0.00} & \textbf{0.01$\pm$0.00} & \textbf{0.01$\pm$0.00} \\
			& Alarm3 & 0.03$\pm$0.01 & \textbf{0.02$\pm$0.00} & 0.05$\pm$0.02 & 0.06$\pm$0.02 & \textbf{0.02$\pm$0.00} & 0.07$\pm$0.02  \\
			& Alarm5 & 0.07$\pm$0.02 & \textbf{0.03$\pm$0.01} & 0.11$\pm$0.04 & 0.09$\pm$0.02 & \textbf{0.03$\pm$0.00} & 0.17$\pm$0.04 \\
			& Alarm10 & 0.16$\pm$0.05 & \textbf{0.10$\pm$0.02} & 0.30$\pm$0.07 & 0.20$\pm$0.05 & \textbf{0.11$\pm$0.02} & 0.56$\pm$0.07 \\
			& Child & \textbf{0.01$\pm$0.00} & \textbf{0.01$\pm$0.00} & \textbf{0.01$\pm$0.00} & \textbf{0.01$\pm$0.00} & \textbf{0.01$\pm$0.00} & \textbf{0.01$\pm$0.00} \\
			& Insurance3 & 0.02$\pm$0.00 & \textbf{0.01$\pm$0.00} & 0.03$\pm$0.01 & 0.03$\pm$0.01 & \textbf{0.02$\pm$0.00} & 0.03$\pm$0.01 \\
\hline
			\hline
		\end{tabular}%

	}
	\label{tab:10}%
	
\end{table*}%

\begin{table*}[htb]
	\centering
	\footnotesize
	\caption{Summary of runtime (in seconds) for HLCD-H and its rivals CMB and ELCS with sample sizes of 500 and 1000}
	{
		\begin{tabular}{cc|ccc|ccc}
			\hline
			\hline
			&&   Size=500 & & &  Size=1000	&\\
			& Network & CMB & ELCS & HLCD-H & CMB & ELCS & HLCD-H \\
			
			\hline
			& Alarm  & 0.04$\pm$0.01 & \textbf{0.01$\pm$0.00} & 0.02$\pm$0.00 & 0.03$\pm$0.00 & \textbf{0.01$\pm$0.00} & 0.02$\pm$0.01 \\
			& Alarm3 & 0.04$\pm$0.01 & \textbf{0.02$\pm$0.01} & 0.05$\pm$0.01 & \textbf{0.05$\pm$0.01} & \textbf{0.05$\pm$0.01} & 0.11$\pm$0.02 \\
			& Alarm5 & 0.06$\pm$0.00 & \textbf{0.04$\pm$0.00} & 0.11$\pm$0.03 & \textbf{0.06$\pm$0.00} & 0.07$\pm$0.01 & 0.17$\pm$0.05 \\
			& Alarm10 & 0.16$\pm$0.01 & \textbf{0.09$\pm$0.01} & 0.50$\pm$0.09 & \textbf{0.13$\pm$0.01} & 0.16$\pm$0.02 & 0.64$\pm$0.10 \\
			& Child & 0.05$\pm$0.01 & \textbf{0.02$\pm$0.01} & 0.03$\pm$0.00 & 0.07$\pm$0.02 & 0.05$\pm$0.01 & \textbf{0.03$\pm$0.01} \\
			& Insurance3 & 0.13$\pm$0.02 & \textbf{0.04$\pm$0.01} & 0.07$\pm$0.01 & 0.22$\pm$0.03 & \textbf{0.05$\pm$0.01} & 0.07$\pm$0.01 \\
\hline
			\hline
		\end{tabular}%

	}
	\label{tab:11}%
	
\end{table*}%

\begin{table*}[tb]
	\centering
	\footnotesize
	\caption{Summary of runtime (in seconds) for HLCD-P and its rival PSL with sample sizes of 500 and 1000}
	{
		\begin{tabular}{cc|cc|cc}
			\hline
			\hline
			&&   Size=500 & &   Size=1000	&\\
			& Network & PSL & HLCD-P & PSL & HLCD-P \\
			
			\hline
			& Alarm & \textbf{0.01$\pm$0.00} & 0.02$\pm$0.00 & \textbf{0.02$\pm$0.00} & \textbf{0.02$\pm$0.00} \\
			& Alarm3 & \textbf{0.04$\pm$0.01} & 0.08$\pm$0.03 & \textbf{0.12$\pm$0.02} & 0.15$\pm$0.02 \\
			& Alarm5 & \textbf{0.09$\pm$0.03} & 0.12$\pm$0.03 & \textbf{0.21$\pm$0.05} & 0.24$\pm$0.05 \\
			& Alarm10 & \textbf{0.46$\pm$0.12} & 0.56$\pm$0.14 & \textbf{0.91$\pm$0.21} & 1.01$\pm$0.22 \\
			& Child & \textbf{0.01$\pm$0.00} & \textbf{0.01$\pm$0.00} & \textbf{0.02$\pm$0.00} & 0.03$\pm$0.00 \\
			& Insurance3 & \textbf{0.04$\pm$0.01} & 0.07$\pm$0.01 & \textbf{0.04$\pm$0.01} & 0.07$\pm$0.01 \\
			& Insurance5 & \textbf{0.08$\pm$0.03} & 0.12$\pm$0.03 & \textbf{0.06$\pm$0.02} & 0.13$\pm$0.04 \\
			& Barley & 10.66$\pm$4.97 & \textbf{1.39$\pm$0.31} & 4.08$\pm$0.26 & \textbf{1.55$\pm$0.16} \\
			& Hailfinder3 & 0.81$\pm$0.15 & \textbf{0.64$\pm$0.12} & \textbf{0.28$\pm$0.03} & 0.56$\pm$0.08 \\
			& Hailfinder5 & 2.36$\pm$0.33 & \textbf{2.26$\pm$0.43} & \textbf{0.53$\pm$0.05} & 1.10$\pm$0.21 \\
			& Hailfinder10 & \textbf{9.89$\pm$1.26} & 12.64$\pm$2.38 & \textbf{1.58$\pm$0.22} & 3.36$\pm$0.50 \\
			& Link & 1.88$\pm$0.43 & \textbf{0.56$\pm$0.16} & 2.80$\pm$0.76 & \textbf{1.95$\pm$0.41} \\
			& Pigs & 0.89$\pm$0.23 & \textbf{0.44$\pm$0.06} & 0.86$\pm$0.15 & \textbf{0.47$\pm$0.08} \\
			& Gene & \textbf{2.01$\pm$0.37} & 2.11$\pm$0.39 & 2.37$\pm$0.68 & \textbf{0.10$\pm$0.02} \\
			\hline
			\hline
		\end{tabular}%

	}
	\label{tab:12}%
	
\end{table*}%

\begin{table*}[htbp]
	\centering
	\caption{Results of ablation experiments for Theorem 1. Total\_PC and Total\_NoPC represent the sum of the number of PC nodes and non-PC nodes for each node in the network, respectively. Get\_PC and Delete\_NoPC refers to the number of PC nodes retained and the number of non-PC nodes removed for each node in the network using Theorem 1.}
	\resizebox{160mm}{!}{
		\begin{tabular}{cc|ccc|ccc}
			\hline
			\hline
			\#Sample & Network & Total\_PC &Get\_PC & Accuracy & Total\_NoPC &Delete\_NoPC & Accuracy \\
			\hline
			\multirow{7}[0]{*}{500} & Alarm5 & 530 & 472.40$\pm$6.65 & 0.89$\pm$0.01 & 33695 & 27300.00$\pm$183.12 & 0.81$\pm$0.01 \\
			& Alarm10 & 1140 & 1002.00$\pm$5.08 & 0.88$\pm$0.00 & 135760 & 108105.80$\pm$577.72 & 0.80$\pm$0.00 \\
			& Hailfinder5 & 916 & 754.00$\pm$13.76 & 0.83$\pm$0.02 & 77484 & 64868.80$\pm$170.30 & 0.84$\pm$0.00 \\
			& Hailfinder10 & 2034 & 1618.60$\pm$16.49 & 0.80$\pm$0.01 & 311566 & 265814.00$\pm$301.27 & 0.85$\pm$0.00 \\
			& Link & 2250 & 1778.40$\pm$0.84 & 0.80$\pm$0.00 & 521926 & 418718.60$\pm$1057.86 & 0.81$\pm$0.00 \\
			& Pigs & 1184 & 1184.00$\pm$0.00 & 1.00$\pm$0.00 & 193297 & 189196.00$\pm$184.21 & 0.98$\pm$0.00 \\
			& Gene & 1944 & 1933.20$\pm$2.86 & 0.99$\pm$0.00 & 639657 & 622343.80$\pm$472.18 & 0.97$\pm$0.00 \\
			\hline
			\multirow{7}[0]{*}{1000} & Alarm5 & 530 & 473.40$\pm$5.34 & 0.89$\pm$0.01 & 33695 & 28930.60$\pm$166.43 & 0.86$\pm$0.00 \\
			& Alarm10 & 1140 & 1002.60$\pm$8.80 & 0.88$\pm$0.01 & 135760 & 116155.00$\pm$454.16 & 0.86$\pm$0.00 \\
			& Hailfinder5 & 916 & 797.20$\pm$8.55 & 0.87$\pm$0.01 & 77484 & 64481.80$\pm$175.08 & 0.83$\pm$0.00 \\
			& Hailfinder10 & 2034 & 1728.80$\pm$14.73 & 0.85$\pm$0.01 & 311566 & 268080.70$\pm$225.01 & 0.86$\pm$0.00 \\
			& Link & 2250 & 1779.00$\pm$1.05 & 0.80$\pm$0.00 & 521926 & 419386.40$\pm$535.96 & 0.81$\pm$0.00 \\
			& Pigs & 1184 & 1184.00$\pm$0.00 & 1.00$\pm$0.00 & 193297 & 187357.80$\pm$132.39 & 0.97$\pm$0.00 \\
			& Gene & 1944 & 1939.00$\pm$1.70 & 1.00$\pm$0.00 & 639657 & 620068.60$\pm$421.87 & 0.97$\pm$0.00 \\
			\hline
			\hline
		\end{tabular}%
	}
	\label{tab:addlabel}%
\end{table*}%

\begin{table*}[htbp]
	\centering
	\caption{Results of ablation experiments for Theorem 2. Total\_V and Total\_NoV respectively denote the total number of V-structures and equivalence class structures in the network. Get\_V and Get\_NoV refer to the correct determination of the number of V-structures and equivalence class structures in the network using Theorem 2.}
	\resizebox{160mm}{!}{
		\begin{tabular}{cc|ccc|ccc}
			\hline
			\hline
			\#Sample & Network & Total\_V & Get\_V & Accuracy & Total\_NoV & Get\_NoV & Accuracy \\		
			\hline	
			\multirow{7}[0]{*}{500} & Alarm5 & 140 & 116.80$\pm$3.99 & 0.83$\pm$0.03 & 635 & 516.20$\pm$7.76 & 0.81$\pm$0.01 \\
			& Alarm10 & 298 & 240.60$\pm$1.84 & 0.81$\pm$0.01 & 1526 & 1245.90$\pm$16.56 & 0.82$\pm$0.01 \\
			& Hailfinder5 & 357 & 262.00$\pm$5.25 & 0.73$\pm$0.01 & 1551 & 1236.40$\pm$11.98 & 0.80$\pm$0.01 \\
			& Hailfinder10 & 850 & 610.40$\pm$9.61 & 0.72$\pm$0.01 & 3767 & 2981.20$\pm$27.17 & 0.79$\pm$0.01 \\
			& Link & 821 & 705.10$\pm$35.34 & 0.86$\pm$0.04 & 6986 & 5821.40$\pm$160.61 & 0.83$\pm$0.02 \\
			& Pigs & 296 & 295.90$\pm$0.32 & 1.00$\pm$0.00 & 3229 & 3228.60$\pm$0.70 & 1.00$\pm$0.00 \\
			& Gene & 269 & 254.90$\pm$1.97 & 0.95$\pm$0.01 & 2691 & 2679.10$\pm$3.35 & 1.00$\pm$0.00 \\
			\hline
			\multirow{7}[0]{*}{1000} & Alarm5 & 140 & 120.30$\pm$2.54 & 0.86$\pm$0.02 & 635 & 523.90$\pm$9.59 & 0.83$\pm$0.02 \\
			& Alarm10 & 298 & 252.10$\pm$2.38 & 0.85$\pm$0.01 & 1526 & 1284.20$\pm$12.97 & 0.84$\pm$0.01 \\
			& Hailfinder5 & 357 & 281.80$\pm$4.66 & 0.79$\pm$0.01 & 1551 & 1279.50$\pm$14.35 & 0.82$\pm$0.01 \\
			& Hailfinder10 & 850 & 671.50$\pm$7.41 & 0.79$\pm$0.01 & 3767 & 3007.20$\pm$16.98 & 0.80$\pm$0.00 \\
			& Link & 821 & 773.80$\pm$15.89 & 0.94$\pm$0.02 & 6986 & 5838.00$\pm$98.12 & 0.84$\pm$0.01 \\
			& Pigs & 296 & 296.00$\pm$0.00 & 1.00$\pm$0.00 & 3229 & 3229.00$\pm$0.00 & 1.00$\pm$0.00 \\
			& Gene & 269 & 257.30$\pm$1.95 & 0.96$\pm$0.01 & 2691 & 2687.20$\pm$1.03 & 1.00$\pm$0.00 \\
			\hline
			\hline
		\end{tabular}%
	}
	\label{tab:addlabel}%
\end{table*}%

\end{document}